\documentclass[lettersize,journal]{IEEEtran}

\usepackage{amsmath,amsfonts}
\usepackage{algorithmic}
\usepackage{algorithm}
\usepackage{array}

\ifCLASSOPTIONcompsoc \usepackage[caption=false,font=normalsize,labelfont=sf,textfont=sf]{subfig}
\else
\usepackage[caption=false,font=footnotesize]{subfig}
\fi
\usepackage{setspace}
\usepackage{textcomp}
\usepackage{stfloats}
\usepackage{amssymb}
\usepackage{url}
\usepackage{verbatim}
\usepackage{multirow} 
\usepackage{graphicx}
\usepackage{cite}
\usepackage{multirow}
\usepackage[table,xcdraw]{xcolor}
\usepackage{colortbl}
\usepackage[normalem]{ulem}
\usepackage{bm}
\usepackage{times}
\usepackage{soul}
\usepackage{url}
\usepackage{amsmath}
\usepackage{amsthm}
\usepackage{booktabs}
\usepackage{algorithm}
\usepackage{algorithmic}
\usepackage[switch]{lineno}
\usepackage{newfloat}
\usepackage{listings}
\usepackage{bibentry}
\usepackage{subcaption}
\usepackage{subfig}
\usepackage{amsfonts}
\usepackage{adjustbox}
\usepackage{soul}
\newtheorem{remark}{Remark}
\usepackage{hyperref}

\begin{document}

\title{Multi-Modal Molecular Representation Learning via Structure Awareness}

\author{Rong Yin, Ruyue Liu*, Xiaoshuai Hao, Xingrui Zhou, Yong Liu, Can Ma, and Weiping Wang
\IEEEcompsocitemizethanks{\IEEEcompsocthanksitem Rong Yin, Ruyue Liu, Can Ma, and Weiping Wang are with the Institute of Information Engineering, Chinese Academy of Sciences, Beijing 100085, China. E-mail: \{yinrong, liuruyue, macan, wangweiping\}@iie.ac.cn.
\IEEEcompsocthanksitem Ruyue Liu is with School of Cyberspace Security, University of Chinese Academy of Sciences, Beijing 100049, China.
\IEEEcompsocthanksitem Xiaoshuai Hao is with Beijing Academy of Artificial Intelligence, Beijing 10086, China. E-mail: xshao@baai.ac.cn.
\IEEEcompsocthanksitem Xingrui Zhou is with Xidian University, Xi'an 710126, China. E-mail: 22009100548@stu.xidian.edu.cn.
\IEEEcompsocthanksitem Yong Liu is with the Renmin University of China, Beijing 100872, China. E-mail: liuyonggsai@ruc.edu.cn.
\IEEEcompsocthanksitem Ruyue Liu is the corresponding author.}}



\IEEEtitleabstractindextext{
\begin{abstract}
Accurate extraction of molecular representations is a critical step in the drug discovery process. In recent years, significant progress has been made in molecular representation learning methods, among which multi-modal molecular representation methods based on images, and 2D/3D topologies have become increasingly mainstream. However, existing these multi-modal approaches often directly fuse information from different modalities, overlooking the potential of intermodal interactions and failing to adequately capture the complex higher-order relationships and invariant features between molecules. To overcome these challenges, we propose a structure-awareness-based multi-modal self-supervised molecular representation pre-training framework (\textbf{MMSA}) designed to enhance molecular graph representations by leveraging invariant knowledge between molecules. The framework consists of two main modules: the multi-modal molecular representation learning module and the structure-awareness module. The multi-modal molecular representation learning module collaboratively processes information from different modalities of the same molecule to overcome intermodal differences and generate a unified molecular embedding. Subsequently, the structure-awareness module enhances the molecular representation by constructing a hypergraph structure to model higher-order correlations between molecules. This module also introduces a memory mechanism for storing typical molecular representations, aligning them with memory anchors in the memory bank to integrate invariant knowledge, thereby improving the model’s generalization ability. 
Compared to existing multi-modal approaches, MMSA can be seamlessly integrated with any graph-based method and supports multiple molecular data modalities, ensuring both versatility and compatibility. Extensive experiments have demonstrated the effectiveness of MMSA, which achieves state-of-the-art performance on the MoleculeNet benchmark, with average ROC-AUC improvements ranging from 1.8\% to 9.6\% over baseline methods.

\end{abstract}

\begin{IEEEkeywords}
Molecular Representation Learning, Self-Supervised, Multi-Modal, Graph Neural Networks.
\end{IEEEkeywords}}

\maketitle
\IEEEdisplaynontitleabstractindextext
\IEEEpeerreviewmaketitle
\ifCLASSOPTIONcompsoc
\IEEEraisesectionheading{\section{Introduction}\label{sec:introduction}}
\else
\section{Introduction}
\label{sec:introduction}
\fi
\IEEEPARstart{I}{n} recent years, the application of machine learning techniques has brought revolutionary changes to the field of drug discovery, especially through significant advancements in molecular image analysis \cite{10424694,liu2025gcl}. While these breakthroughs have propelled research forward, a fundamental challenge remains: how to effectively generate vectorized embeddings that capture the complex structure and functional characteristics of molecules. With the maturation of supervised learning \cite{yin2019sketch,yin2020divide,rong2020self,yin2022distributed}, unsupervised learning has attracted growing research interest \cite{yin2022randomized,yin2022scalable,yin2020extremely}. Self-supervised learning \cite{hao2025mapfusion, liu2023aswt,10308718,hao2025bctr} has emerged as a promising solution to this problem, as it allows for pre-training on large datasets and fine-tuning specific tasks to achieve exceptional performance. This approach has garnered widespread attention. However, a key challenge in self-supervised pre-training is designing a rich latent space for molecules. This requires not only the construction of an appropriate encoder but also the formulation of effective training objectives to guide the learning process.

Molecules can be directly represented as images through electron microscopy, X-ray crystallography, or molecular rendering tools, allowing for visualization of their shapes and internal structures \cite{shen2022atomic, landrum2013rdkit,661006}. These images provide intuitive geometric information for molecular modeling, enabling the capture of spatial features of molecules. Additionally, graph-based molecular representations, where atoms are nodes and chemical bonds are edges, offer the most natural and direct representation of molecular structures, making them a core method for molecular modeling. Graph representations accurately reflect molecular structures and effectively reveal the chemical properties and biological activities of molecules during the drug discovery process \cite{hu2020strategies, 10539072}. Consequently, graph-based molecular representations have been extensively explored and applied in drug discovery, molecular optimization, and bioinformatics \cite{jiang2022molecular, xia2023mole}.
Existing self-supervised pre-training methods can be classified into four main types: (1) \textbf{2D graph masking} \cite{hu2020strategies, rong2020self, hou2022graphmae,10179964}, where random portions of a molecular graph are masked, and the model is trained to recover them; (2) \textbf{3D graph denoising} \cite{zaidi2022pre, feng2023fractional, liu2023molecular}, where noise is added to 3D molecular conformations, and the model is trained to predict the noise; (3) \textbf{molecular image generation} \cite{jiang2022molecular,tchagang2021time}, which extracts critical features from the visual information of molecules; and (4) \textbf{multi-modal methods} \cite{liu2021pre,karim2021quantitative}, which combine 2D and 3D graph information for contrastive learning by aligning the representation of 3D conformations with their corresponding 2D graphs. Although these methods enhance molecular representation, they still have significant limitations, leaving room for further improvement: (1) The representation and encoding strategies for 2D and 3D graphs are often similar, leading to insufficient complementary information between the two modalities \cite{wu2020comprehensive}; (2) Many existing methods focus on modality fusion but fail to capture complex higher-order relationships and invariant features within molecules, particularly the nonlinear dependencies between different molecular groups, which are critical for molecular properties and activities.

To overcome these challenges, we propose a structure-awareness-based multi-modal self-supervised molecular representation learning pre-training framework (\textbf{MMSA}). The framework consists of two main components: the multi-modal molecular representation learning module and the structure-awareness module. The multi-modal molecular representation learning module employs multiple auto-encoders to learn latent representations from different modalities. Since molecular images can effectively capture texture information and visualize molecular structure without relying on conformations, we incorporate image-derived prior knowledge to aid representation learning. The structure-awareness module models higher-order relationships between molecules by constructing a hypergraph structure, capturing complex dependencies between components, and enhancing molecular representations.
Additionally, this module introduces a memory mechanism that stores typical molecular representations in a memory bank and aligns them with memory anchors to integrate invariant knowledge, significantly boosting the model's generalization ability. This ensures strong prediction performance when the model encounters new molecular data.

The main contributions of this paper are as follows:
\begin{itemize}
    \item We propose a multi-modal self-supervised pre-training framework (\textbf{MMSA}) for molecular representation learning, which combines a multi-modal molecular representation learning module and structure-awareness module to enhance molecular representations and capture higher-order relationships between molecules.
    \item Compared to existing multi-modal learning methods, the proposed MMSA can be integrated with any graph-based learning method and supports various molecular data modalities, not limited to graphs or images. It demonstrates versatility at the model level and compatibility at the data level.
    \item We introduce hypergraph structures to model higher-order relationships in molecular representations and incorporate a memory mechanism to integrate invariant molecular knowledge, thus improving generalization. To our knowledge, we are the first to model higher-order relationships in multi-modal molecular representation learning using hypergraphs.
    \item Extensive experiments demonstrate that the proposed method outperforms existing methods across three downstream tasks (classification, regression, and retrieval) on MoleculeNet benchmarks, significantly enhancing the performance of both graph-based molecular models.
\end{itemize}

The paper is organized as follows: Section \ref{Related Work} provides an overview of related work on molecular representation learning. Section \ref{Preliminary} introduces the preliminaries used in the paper. Section \ref{Method} presents the details of the proposed method. Section \ref{Experiment} describes the experiments to evaluate the proposed method. Finally, We conclude this paper in Section \ref{conclusion}.

\section{Related Work}
\label{Related Work}
\subsection{Graph-based Molecular Representation Learning}
Graph data offers a distinct advantage in efficiently storing and representing complex structural information, making Graph Neural Networks (GNNs) a dominant approach for tasks such as molecular property prediction and drug discovery. However, annotating molecular data for training GNN models is a high-cost process, which has led recent research to focus on leveraging pre-training strategies with large-scale, unlabelled molecular datasets. For example, GPT-GNN \cite{hu2020gpt} introduces a property graph generation task that facilitates the pre-training of GNNs to capture better both the structural and semantic properties of molecular graphs. Hu et al. \cite{hu2020strategies} and Li et al. \cite{li2021effective} explore property and structural prediction on molecular graphs, focusing on node-level and graph-level tasks, respectively. These methods aim to utilize the intrinsic properties of graphs to enhance the accuracy of molecular property prediction. To capture the complex relationships within molecular structures, approaches such as GROVER \cite{rong2020self}, and MGSSL \cite{zhang2021motif} propose predicting or generating molecular graph templates, thereby improving the model's ability to learn molecular graph patterns. Mole-BERT \cite{xia2023mole} introduces a novel approach with mask-atom modeling and ternary mask-contrast learning, optimizing mask-based GNN architectures. This technique enables GNNs to focus on specific molecular regions, allowing for more granular learning of molecular properties. Given the critical role of 3D geometric information in accurately predicting molecular properties, several recent studies \cite{liu2021pre,zhu2022unified,stark20223d} have extended GNN pre-training to incorporate 3D geometric data, which enriches the representation of molecular structures and improves predictive accuracy. We recommend the recent surveys \cite{guo2022graph, fang2022geometry}, which provide an in-depth review of the research in this area. In this paper, we follow the graph-based paradigm for molecular property prediction and aim to enhance and optimize existing graph-based models through novel approaches.

\subsection{Image-based Molecular Representation Learning}
The remarkable success of deep learning methods in computer vision has inspired new approaches to molecular representation learning. MolMap \cite{tchagang2021time} maps molecular descriptors and fingerprint features into 2D feature maps to capture the intrinsic relationships between molecular characteristics. MolPSI \cite{jiang2022molecular} represents molecules as 2D images of uniform size based on spectral models. With the growing adoption of self-supervised learning in computer vision, the first pre-trained model based on molecular images, known as ImageMol \cite{zeng2022accurate}, was introduced. This model learns representations from 10 million molecules and is trained with five carefully designed pre-training tasks.
Additionally, molecular images have been used to detect functional groups, with the pre-trained model being further fine-tuned for molecular activity prediction \cite{iqbal2021learning}. However, image-based methods have not become the dominant approach in molecular representation learning because they require transforming data samples into Euclidean space, which fails to accurately represent atomic and bond attributes, making them unsuitable for direct molecular property prediction. Despite these limitations, image-based methods still promise to complement molecular representation learning in specific contexts, particularly for visual perception tasks. In this paper, we expand the use of images in the molecular domain by incorporating image features and self-supervised learning strategies to enhance model performance and predictive accuracy.

\subsection{Multi-Modal Molecular Representation Learning}
Various molecular representations, including 1D (e.g., SMILES), 2D, and 3D formats, employ different strategies to capture molecular structures. Combining these representations offers a multi-perspective view of the molecule, providing a more comprehensive understanding of its structure and properties. The GraSeq method \cite{guo2020graseq} enhances molecular representation by integrating molecular maps and SMILES sequences, encoding them with graph convolutional networks (GCNs) \cite{kipf2016semi} and bi-directional long short-term memory networks (biLSTMs) \cite{huang2015bidirectional}, respectively, to capitalize on the strengths of both representations. Karim et al. \cite{karim2021quantitative} combine SMILES, fingerprint maps, molecular maps, and 2D and 3D descriptors with various deep learning (DL) models for quantitative toxicity prediction. GraphMVP \cite{liu2021pre} further advances this approach by combining 2D and 3D graphs within a contrastive learning framework to train the model. IME \cite{xiangimage} incorporates the rich visual information embedded in 3D molecular conformations—such as texture, shading, and spatial features—and utilizes cross-modal knowledge distillation to generate more discriminative representations for drug discovery. Although these approaches have made significant progress in multi-modal representation learning, they typically overlook the interactions between modalities when fusing information, limiting the model's ability to capture potential correlations and complementarities. Moreover, most methods fail to fully account for invariant structural information across molecules, which is crucial for accurate molecular property prediction and modeling. Therefore, effectively combining multiple representations, fully exploiting inter-modal interactions, and leveraging invariant structural information remain critical challenges in molecular representation learning.

\begin{table}[t]
\centering
\caption{Summary of key notations.}
\begin{adjustbox}{max width=0.45\textwidth}
\begin{tabular}{ll}
\toprule
\textbf{Symbol} & \textbf{Description} \\
\midrule
$\mathcal{G}_i$ & 2D molecular graph \\
$\mathcal{V}_i$ & The set of nodes (atoms)\\
$\mathcal{E}_i$ & The set of edges (bonds)\\
$\bm{X}_i^v$ & Atomic feature matrix for graph $\mathcal{G}_i$ \\
$\bm{X}_i^e$ & Bond feature matrix for graph $\mathcal{G}_i$ \\
$\bm{I}_{2D}$ & 2D molecular image representation \\
$\mathcal{F}_1, \mathcal{F}_2, \mathcal{F}_3$ & Feature extractors for 2D graph, 2D image, and 3D graph \\
$\bm{c}_i$ & Latent embedding for modality $m_i$ \\
$\bm{Z}$ & Structure-awareness embeddings \\
$\bm{a}_j$ & Memory anchor in memory bank \\
$\lambda, \alpha$ & Hyperparameters balancing loss terms \\
$K$ & Number of nearest neighbors for hypergraph construction \\
\bottomrule
\end{tabular}
\end{adjustbox}
\label{tab:symbols}
\vspace{-1em}
\end{table}

\section{Preliminary}
\label{Preliminary}
We first outline the key concepts and notations employed in this work. Self-supervised molecular representation learning (SSL) relies on the design of multiple views, each capturing distinct aspects or modalities of the data. Each molecule inherently possesses multiple views, denoted as \(\mathcal{M}=\{m_i\}_{i=1}^{M}\). For example, a 2D molecular graph represents the topological structure defined by adjacency relationships, while a 2D image provides a visual representation of the molecule, incorporating texture information and enabling spatial visualization without requiring any conformational assumptions. In contrast, a 3D molecular graph more accurately reflects the geometric and spatial relationships. By integrating these diverse views, SSL methods can leverage the strengths of each representation, leading to more robust and accurate molecular modeling.

\textbf{The 2D molecular graphs} represent molecules astwo-dimensional graphs \(\{\mathcal{G}_i = (\mathcal{V}_i, \mathcal{E}_i)\}_{i=1}^N\), where \(\mathcal{V}_i\) denotes the set of nodes (atoms) and \(\mathcal{E}_i \in \mathcal{V}_i \times \mathcal{V}_i\) represents the set of edges (chemical bonds) of a graph $\mathcal{G}_i$. $N$ denotes the number of molecules. The graph \(\mathcal{G}_i\) is typically associated with an atomic feature matrix \(\bm{X}_i^v \in \mathbb{R}^{n_i^v \times d_i^v}\) and a bond feature matrix \(\bm{X}_i^e \in \mathbb{R}^{n_i^e \times d_i^e}\), where \(n_i^v\) and \(n_i^e\) denote the number of atoms and bonds, respectively, and \(d_i^v\) and \(d_i^e\) represent the dimensionality of atomic and bond features. Given a 2D molecular graph \(\mathcal{G}\), the molecular graph representation is derived using a 2D graph neural network (GNN):
\begin{equation}
\bm{x}_{2D-Graph} = \mathcal{F}_1(\bm{X}^v, \bm{X}^e).
\label{eq1}
\end{equation}

\textbf{The 2D molecular images} corresponding to the molecular graph \( \mathcal{G} \) can be represented as \( \bm{I}_{2D} \in \mathbb{R}^{H \times W \times 3} \), where \( H \) and \( W \) denote the height and width of the image, respectively. In addition to the 2D image, the molecule can also be represented as a multi-view 3D image \( \bm{I}_{3D} \in \mathbb{R}^{V \times H \times W \times 3} \), where \( V \) represents the number of views. However, due to the high computational cost of 3D image training, this paper focuses exclusively on single-view 2D images. Using RDKit \cite{landrum2013rdkit}, a single-view 2D image can be generated directly, and the 2D image representation can be extracted using the corresponding image encoder, as expressed in the following equation:
\begin{equation}
\bm{x}_{2D-Image} = \mathcal{F}_2(\bm{I}_{2D}).
\label{eq2}
\end{equation}

\textbf{The 3D molecular graphs} include not only the properties of the atoms and chemical bonds but also the spatial positions of the atoms, denoted as \( \bm{R} \in\mathbb{R}^{n^{v} \times 3} \). The three-dimensional structure of an atom moving continuously on the potential energy surface, and located at a local minimum on this surface, is referred to as a conformation. A conformation is an isomer of a molecule that is distinguished from another isomer by the rotation of a single bond. Given a conformation \( \mathcal{C} = ( \mathcal{V}, \mathcal{E}, \bm{R}) \), it is represented by a 3D GNN model as follows:
\begin{equation}
\bm{x}_{3D-Graph} = \mathcal{F}_3(\bm{X}^v, \bm{X}^e, \bm{R}).
\label{eq3}
\end{equation}

For simplicity, we henceforth denote the 2D graph representation, 2D image representation, and 3D graph representation as \( \bm{x}_1 \), \( \bm{x}_2 \), and \( \bm{x}_3 \), respectively.

\begin{figure*}
    \centering
    \includegraphics[width=0.9\textwidth]{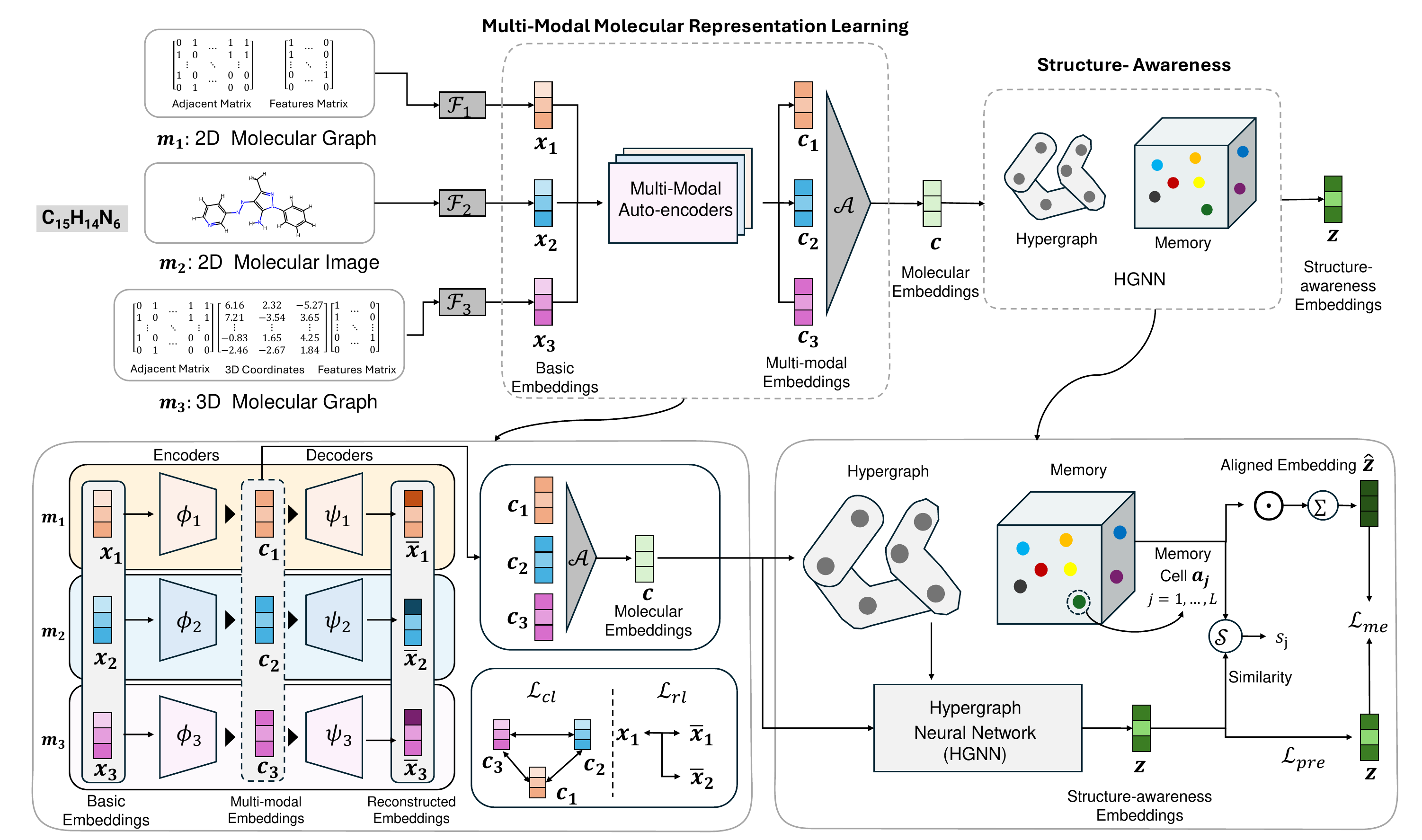}
    \caption{The proposed MMSA pre-training framework consists of two main modules: the multi-modal molecular representation learning module and the structure-awareness module. \(\mathcal{F}_i\) represents the method for learning the feature representation of modality \( m_i \), where \( m_i \) can be 2D graphs, 2D images, 3D graphs, etc. In the multi-modal molecular representation learning module, the molecular embedding \( c \) is generated by aggregating the modality embeddings \( \{ c_i \}_{i=1}^M \). The \( \mathcal{L}_{cl} \) loss is designed to capture shared information across modalities, while the \( \mathcal{L}_{rl} \) loss focuses on enhancing the generalization ability of each modality embedding. In the structure-awareness module, hypergraphs capture higher-order correlations between molecular graphs, enabling the learning of generalizable molecular structures. The memory mechanism is developed to further enhance the generalization ability of multi-modal molecular embeddings by retaining and reusing crucial structural and contextual information, thereby improving the robustness and adaptability of the learned representations.}
    \label{fig1}
    \vspace{-1em}
\end{figure*}

\section{Methodology}
\label{Method}
In this section, we propose the structure-awareness-based multi-modal self-supervised representation learning framework (MMSA) for molecular graphs. The overall architecture of MMSA is shown in Fig. \ref{fig1}, consisting of two main modules: the multi-modal molecular representation learning module and the structure-awareness module. Given a molecule associated with multiple modalities (e.g., 2D graphs, 2D images, 3D graphs), typical feature extraction methods are employed to capture the fundamental characteristics of each modality. The multi-modal molecular representation learning module is then introduced to generate a unified molecular graph embedding from the basic multi-modal semantic features. In the structure-awareness module, hypergraph convolution and a memory bank capture higher-order correlations and invariant knowledge between molecules. Finally, the pre-trained model can be fine-tuned for downstream tasks such as classification, regression, and retrieval.
\subsection{Multi-Modal Molecular Representation Learning}
Constrained by the limitations of uni-modal representations, we extract unified molecular embeddings from multi-modal representations. Specifically, we first apply typical feature representation methods to extract the basic features of each modality. These features are then mapped to the latent space of each modality using an auto-encoder. During this process, a contrastive loss is employed to bring the encoded representations of different modalities closer together, ensuring that modalities from the same object are more similar than those from other objects. Additionally, to minimize information loss during compression, we introduce both intra-modal and cross-modal reconstruction losses.
\subsubsection{Multi-Modal Auto-Encoders}
Given a molecule and \( M \) feature extractors \( \{\mathcal{F}_i \}_{i=1}^{M} \), we generate the basic feature matrices \( \{ \bm{x}_i \}_{i=1}^{M} \), where \( \bm{x}_i \in \mathbb{R}^{d_o} \), and $d_o$ is the output dimension. Specifically, we use the Graph Isomorphism Network (GIN) \cite{xu2018powerful} to obtain a 2D graph-based representation of the molecule, ResNet-18 \cite{he2016deep} to generate a 2D image-based representation, and ComENet \cite{wang2022comenet} to capture a 3D graph-based representation of the molecule. As shown in Fig. 1, the auto-encoder compresses the input features of modality \( \bm{x}_i \) into the latent space \( \bm{c}_i \), which represents the molecular embedding space, to obtain improved representations. This process is defined as follows:
\begin{equation}
    \left.\left\{\begin{matrix}\text{Encoder}: \phi_i=\bm{x}_i\to \bm{c}_i\\\text{Decoder}: \psi_i=\bm{c}_i\to \bar{\bm{x}}_i\end{matrix}\right.\right.,
\label{eq4}
\end{equation}
where \( \bm{c}_{i} = \phi_i(\bm{x}_{i}) \) represents the molecular embedding estimated for modality \( m_i \), and \( \bm{c}_{i} \in \mathbb{R}^{d_c} \). The modality feature reconstructed from \( \bm{c}_{i} \) can be expressed as \( \bar{\bm{x}}_i = \psi_i(\bm{c}_{i}) \), and \( \bar{\bm{x}}_i \in \mathbb{R}^{d_o} \).

\subsubsection{Optimization Objectives}
To leverage the complementary information across modalities, we introduce two loss functions: the contrastive loss and the reconstruction loss. 

\textbf{Contrastive Loss:} 
The contrastive loss is designed to minimize the distance between the embeddings \( \{ \bm{c}_{i} \}_{i=1}^{M} \) of the same molecule across different modalities. Specifically, the contrastive loss encourages the embeddings of the same molecule from different modalities to be closer in the latent space, promoting a more cohesive representation of the molecule across modalities. It is defined as follows:
\begin{equation}
\begin{aligned}
\mathcal{L}_{cl} = &-\frac{2}{M(M+1)} \times   \\ &\sum_{i=1}^{M}\sum_{j\ne i}^M 
\log \frac{e^{\theta\left(\bm{c}_i, \bm{c}_j\right) }}{e^{\theta\left(\bm{c}_i, \bm{c}_j\right) }+ e^{\theta\left(\bm{c}_i, \bar{\bm{c}}_i\right) }+e^{\theta\left(\bm{c}_i, \bar{\bm{c}}_j\right)}},
\label{eq5}
\end{aligned}
\end{equation}
where \( \theta(\cdot) \) denotes the cosine similarity function, and \( \bm{c}_i \) and \( \bm{c}_j \) represent the estimated molecular embeddings from the same molecule but with different models (i.e., positive pairs). \( \bar{\bm{c}}_i \) and \( \bar{\bm{c}}_j \) denote negative samples, representing the estimated embeddings of different molecules within the same batch. The contrastive loss \( \mathcal{L}_{cl} \) ensures that embeddings from the same molecule across distinct modalities are pulled closer together in the latent space. This encourages a more compact and aligned multi-modal representation, improving the coherence and robustness of the learned embeddings across modalities.

\textbf{Reconstruction Loss:} 
To enhance the generalization ability of the encoders \( \{ \phi_i(\cdot) \}_{i=1}^M \) and decoders \( \{ \psi_i(\cdot) \}_{i=1}^M \) for different molecular modalities, we propose the reconstruction loss. The Reconstruction Loss consists of two components: intra-modal reconstruction loss and cross-modal reconstruction loss. The loss function is formulated as follows:

\begin{equation}
\begin{aligned}
\mathcal{L}_{rl}=&\frac{1}{M}\sum_{i=1}^{M}\sum_{j\neq i}^M\Big(\tau\left\|\bm{x}_{i}-\bar{\bm{x}}_{i}\right\|_{2}\\ &+(1-\tau)\|\bm{x}_{i}-\psi_{i}(\phi_{j}(\bm{x}_{j}))\|_{2}\Big),
\label{eq6}
\end{aligned}
\end{equation}
where \( \bm{x}_i \) and \( \bar{\bm{x}}_i \) denote the basic features and the reconstructed features of the molecule for modality \( m_i \), respectively. The first term in Eq. \eqref{eq6} represents the intra-modal reconstruction, ensuring that features of each modality are accurately reconstructed within the same modality. The second term corresponds to the cross-modal reconstruction, where \( i \neq j \) forces the model to preserve cross-modal consistency by enforcing that features from different modalities are reconstructed relative to each other. Here, \( \tau \) is a hyperparameter used to balance the reconstruction ability of the auto-encoders with their generalization capacity.

By incorporating intra-modal and cross-modal reconstruction constraints, this loss function strengthens the ability of encoder-decoder pairs to reconstruct features within each modality. It ensures that the representations across modalities remain consistent. This dual objective promotes both the accuracy and the generalization of the model, improving its robustness when handling diverse molecular data.

\textbf{Overall Loss:} 
By combining the contrastive loss from Eq. \eqref{eq5} and the reconstruction loss from Eq. \eqref{eq6}, we define the overall loss function for multi-modal molecular representation learning as:
\begin{equation}
\mathcal{L}_{ae} = \lambda \mathcal{L}_{cl} + (1 - \lambda) \mathcal{L}_{rl},
\label{eq7}
\end{equation}
where \( \lambda \) is a hyperparameter that controls the trade-off between the contrastive loss \( \mathcal{L}_{cl} \) and the reconstruction loss \( \mathcal{L}_{rl} \).

Finally, the model generates molecular embeddings \( \{ \bm{c}_i \}_{i=1}^{M} \), with each embedding being estimated from a different modality. Since these embeddings are mapped to the same latent space, we introduce an aggregation function \( \mathcal{A}(\cdot) \) to generate a unified multi-modal molecular embedding, expressed as:
\begin{equation}
\bm{c} = \mathcal{A}\left( \{ \bm{c}_i \}_{i=1}^{M} \right),  
\end{equation}
where \( \bm{c} \) represents the unified molecular embedding that integrates information from all modalities. This aggregation step consolidates the diverse representations into a single, comprehensive embedding, capturing the full spectrum of molecular features and enhancing the ability of models to learn from multi-modal data.

\subsection{Structure-Awareness}
Although existing multi-modal molecular representation learning methods have improved the accuracy of molecular representations, they still need to capture the complex higher-order relationships and invariant features between molecules. The higher-order correlations of molecular structures, especially the nonlinear dependencies between different atoms and functional groups, often influence their properties and activity. To further enhance the representational capacity of molecules, we propose a structure-awareness module that models the higher-order correlations between molecules by constructing a hypergraph structure. This hypergraph structure effectively captures the intricate relationships among various components within the molecule, thereby improving the molecular representation. Moreover, the structure-awareness module introduces a memory mechanism that stores typical molecule representations in a memory bank. By aligning with the memory anchor points in this bank, the model integrates invariant knowledge from the molecule, allowing it to maintain strong predictive performance even when confronted with new molecular data.
\begin{figure}
    \centering
    \includegraphics[width=0.8\linewidth]{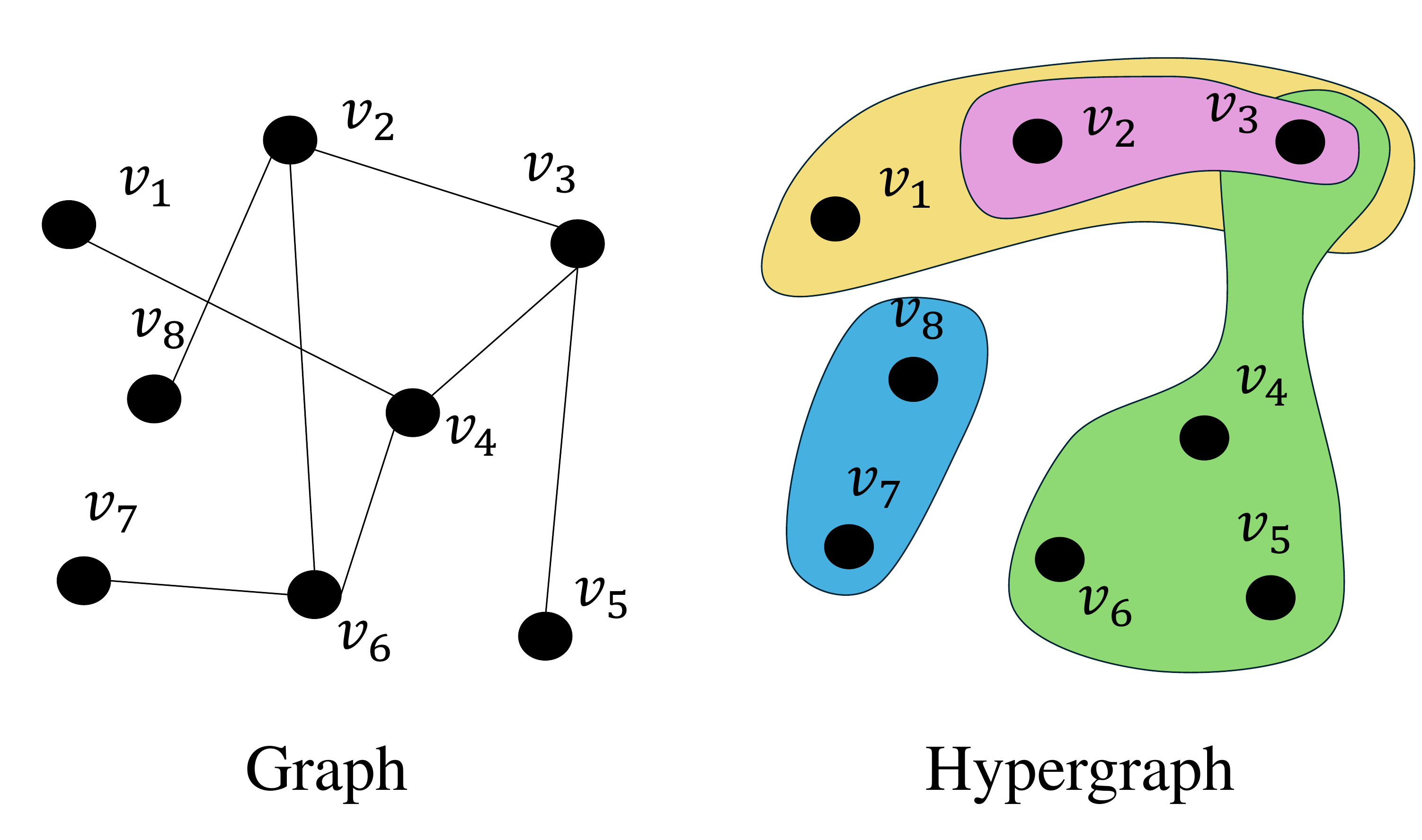}
    \caption{Comparison of Graph and Hypergraph.}
    \label{fig2}
    \vspace{-1em}
\end{figure}
\subsubsection{Hypergraph Network}
As shown in Fig. \ref{fig2}, in traditional graphs, each node \(v\) represents a molecule, while edges denote direct relationships or interactions between molecules. However, this structure can only capture pairwise, binary interaction information. When molecules exhibit shared, complex interactions or common properties, traditional graphs are limited in their ability to represent multivariate interactions through simple edge connections. In contrast, hypergraphs introduce the concept of hyperedges, which can simultaneously connect multiple nodes, thereby capturing higher-order dependencies between molecules. This enhances the model's ability to understand complex molecular systems and allows for better handling of intricate interactions between multiple molecules. A hypergraph can be represented as \(\mathcal{G}_h = \{ \mathcal{V}_h, \mathcal{E}_h, \bm{A}_h, \bm{H}_h \}\), where \(\mathcal{V}_h\) and \(\mathcal{E}_h\) are the sets of nodes and hyperedges, respectively. The hypergraph adjacency matrix \(\bm{H}_h \in \{0, 1\}^{|\mathcal{V}_h| \times |\mathcal{E}_h|}\), where the \(i\)-th hyperedge corresponds to the \(i\)-th column of \(\bm{H}_h\), and \(\bm{H}_h(v, e) = 1\) if hyperedge \(e\) contains node \(v\). Additionally, \(\bm{H}_h \in \mathbb{R}^{|\mathcal{E}_h| \times |\mathcal{E}_h|}\) is a diagonal matrix, where \(\bm{A}(i,i)\) represents the weight of the \(i\)-th hyperedge.

\textbf{Hypergraph Convolution:} 
In the proposed method, each molecule with a multi-modal representation is treated as a node. Through the multi-modal molecular representation learning module, a unified molecular embedding matrix \( \bm{C} = \{\bm{c}^l\}_{l=1}^N \) is generated \footnote{For simplicity, in Section 4.1, we omit the superscript $l$ that denotes the molecular index.}, where \( \bm{C} \in \mathbb{R}^{N \times d_c} \), and this matrix serves as the feature matrix associated with each node. Next, hyperedges are constructed using the \( K \)-nearest neighbors (KNN) algorithm to model the high-order, synergistic information between molecules. Specifically, for each node, a hyperedge is created by connecting it to its $K-1$ nearest neighbor nodes, constructing the \( N \) hyperedges. To learn structure-awareness embeddings \( \bm{Z} \in \mathbb{R}^{N \times d_c} \) from the hypergraph, we apply hypergraph convolution (HGNNConv) \cite{gao2022hgnn+} as follows: 
\begin{equation}
\bm{Z}=\sigma\left(\bm{D}_v^{-\frac{1}{2}}\bm{A_h}\bm{H}_h\bm{D}_e^{-1}\bm{A}_h^\top\bm{D}_v^{-\frac{1}{2}}\bm{C}\bm{W}\right),
\end{equation}
where \( \bm{D}_v \) and \( \bm{D}_e \) are the diagonal degree matrices for nodes and hyperedges, respectively. \( \bm{W} \in \mathbb{R}^{d_c \times d_c} \) is the learnable parameter of the HGNNConv layer, and $\sigma(\cdot)$ is activation function (ReLU in our implementation). Applying HGNNConv across the hypergraph constructed from all molecules, the resulting structure-awareness embeddings \( \bm{Z} \) effectively capture the complex higher-order interactions between molecules. This enables the model to understand better and leverage the intricate relationships within the molecular system.

\textbf{Memory Bank:}
In addition to modeling the relationships between molecules using the hypergraph structure, we also attempt to extract representative features from the molecules. To capture invariant information, we introduce a memory bank that stores many typical representations during the training phase. The memory bank consists of \( L \) invariant memory anchors \( \bm{a}_j \), where each anchor can store a typical representation of a molecule. This can be expressed as \( \{ \bm{a}_j \in \mathbb{R}^{d_c} \}_{j=1}^L \).

Given the structure-awareness embedding \( \bm{z} \) of a molecule,\footnote{Here, \( \bm{z} \) denotes the structure-awareness embedding of a molecule. For simplicity, we omit superscripts indicating the molecule index.} we first compute the activation score \( s_{j} \) for each memory anchor \( \bm{a}_j \) in the memory bank. These activation scores reflect the similarity between the molecular embedding and the prototypical representations stored in the memory. Specifically, each memory anchor \( \bm{a}_j \in \mathbb{R}^d \) is initialized from a uniform distribution, following the standard initialization strategy for embedding vectors to ensure variance stability. During training, the memory anchors are continuously refined through attention-based interactions with molecular embeddings and are optimized jointly with the rest of the model parameters. It is defined as follows:
\begin{equation}
    s_{j} = \mathcal{S}(\bm{z}, \bm{a}_j),
\end{equation}
where \( \mathcal{S}(\cdot, \cdot) \) is a distance metric function. Then, we apply the softmax function to normalize the \( L \) activation scores \( s_{j} \). The normalized activation scores \( s'_{j} \) are used to reconstruct the molecular embedding from all the memory anchors in the memory bank, as follows:
\begin{equation}
\begin{aligned}
s'_{j} &= \frac{e^{s_{j}}}{\sum_{j=1}^{L} e^{s_{j}}}, \quad 
\hat{\bm{z}} &= \sum_{j=1}^{L} s'_{j} \bm{a}_j,
\end{aligned} 
\end{equation}
where \( \hat{\bm{z}} \in \mathbb{R}^{d_c} \) represents the aligned molecular embedding. The memory bank stores invariant and meaningful semantic knowledge of molecular objects. By aligning the embedding to typical memory anchors, we can mitigate overfitting issues and learn invariant knowledge that generalizes better across different molecular data. This approach helps the model focus on the essential, consistent features of the molecular system, improving its robustness and predictive power.

\subsubsection{Optimization Objectives}
To train the hypergraph convolution and the learnable memory anchors, we adopt two loss functions: the memory loss \( \mathcal{L}_{me} \) and the common prediction loss \( \mathcal{L}_{pre} \).

\textbf{Memory Loss}: 
Memory loss ensures that the structure-awareness and aligned embeddings remain similar. It also updates the relevant memory anchors to capture invariant and typical knowledge about the generalized molecular representations. Specifically, the memory loss is defined as:
\begin{equation}
 \mathcal{L}_{me} = \| \hat{\bm{z}} - \bm{z} \|^2   
\end{equation}
where \( \bm{z} \) is the original structure-awareness molecular embedding, and \( \hat{\bm{z}}\) is the aligned embedding. By minimizing the loss, the model can retain the structural information in the embeddings and refine the memory anchors to represent better the essential and invariant characteristics of the molecular system, aiding in generalization.

\textbf{Prediction Loss:}
Unlike most methods that rely on node-level information (such as atoms) or edge-level features (such as bonds), our goal is to extract fundamental prior knowledge from two graph-level attributes to guide the learning of molecular representations: molecular geometry \( \bm{Y}^{geom} \) and chemical properties \( \bm{Y}^{prop} \). Specifically, we input the structure-awareness embeddings and the aligned embeddings into two separate predictors: a geometry predictor and a chemical property predictor. Both predictors share the same architecture: fully connected layers (FC) → Softplus activation → fully connected layers (FC). The prediction loss is defined as:
\begin{equation}
\begin{aligned}
\mathcal{L}_{pre} = &| \bm{Y}^{geom} - \hat{\bm{Y}}^{geom}_{\bm{z}} | + | \bm{Y}^{prop} - \hat{\bm{Y}}^{prop}_{\bm{z}} | \\ &+ | \bm{Y}^{geom} - \hat{\bm{Y}}_{\hat{\bm{z}}}^{geom} | + | \bm{Y}^{prop} - \hat{\bm{Y}}_{\hat{\bm{z}}}^{prop}|, 
\end{aligned}
\end{equation}
where \( \bm{Y}_{geom} \) and \( \bm{Y}_{prop} \) represent the true labels for molecular geometry and chemical properties, respectively, and \( \hat{\bm{Y}}^{geom}_{\bm{z}} \) and \( \hat{\bm{Y}}^{prop}_{\bm{z}} \) are the predicted values generated by the geometry and property predictors using the structure-awareness embeddings \( \bm{z} \). Similarly, \( \hat{\bm{Y}}_{\hat{\bm{z}}}^{geom} \) and \( \hat{\bm{Y}}_{\hat{\bm{z}}}^{prop} \) are the predicted values generated using the aligned embeddings \( \hat{\bm{z}} \). This prediction loss ensures the model effectively maps molecular embeddings to their corresponding geometric and chemical features, providing a more comprehensive understanding of the molecular.

In the structure-awareness module, to effectively combine the memory loss \( \mathcal{L}_{me} \) and the prediction loss \( \mathcal{L}_{pre} \), we define a weighted loss function:
\begin{equation}
\mathcal{L}_{sa} = \alpha \mathcal{L}_{me} + (1 - \alpha) \mathcal{L}_{pre} , 
\end{equation}
where \( \alpha \) is a hyperparameter that controls the balance between the memory loss \( \mathcal{L}_{me} \) and the prediction loss \( \mathcal{L}_{pre} \).

Finally, the multi-modal molecular model can be pre-trained using the following total loss function:
\begin{equation}
\mathcal{L}_{overall} = \mathcal{L}_{ae} + \mathcal{L}_{sa}.
\end{equation}

This total loss function integrates the objectives of both multi-modal representation learning and structure-awareness learning, ensuring that the model learns a unified and effective multi-modal embedding and captures high-order relationships and invariant knowledge within the molecular system. 

In summary, compared to existing multi-modal methods, the proposed MMSA framework offers significant advantages in multi-modal molecular representation learning. While current approaches typically rely on modality fusion, they often fail to effectively capture the complex higher-order relationships and invariant features of molecules, limiting the quality of molecular representations and hindering the full exploration of their structural and functional properties. MMSA addresses these limitations by integrating a multi-modal molecular representation learning module with a structure-awareness module, enhancing the model's ability to capture complex molecular structures and higher-order relationships, thus improving the expressiveness of molecular representations. Moreover, the incorporation of hypergraph structures and memory mechanisms enables the effective integration of invariant molecular knowledge, greatly enhancing generalization and robustness across various tasks. Furthermore, MMSA is highly versatile and compatible, seamlessly integrating with any graph-based learning method and supporting multiple molecular data modalities, including graphs, images, and more, highlighting its broad application potential and flexibility.

\begin{remark}
The time complexity of constructing a hypergraph is \( \mathcal{O}(N^2 d_c) \). Given the reconstructed embeddings of $N$ molecules, \( \bm{C} \in \mathbb{R}^{N \times d_c} \), and a hyperparameter \( K \), the hypergraph is constructed using the \( K \)-nearest neighbors (KNN) algorithm, which involves three main steps: (1) computing the distance matrix, (2) sorting the neighbors, and (3) constructing the hypergraph adjacency matrix. First, if the distance between two objects is measured using the inner product, the time complexity for computing the distance matrix is \( \mathcal{O}(N^2 d_c) \). Second, the time complexity for sorting the neighbors is \( \mathcal{O}(N^2 \log N) \). The third step, constructing the hypergraph adjacency matrix based on the nearest neighbors, has a time complexity of \( \mathcal{O}(NK) \). 
Thus, the overall time complexity of constructing a hypergraph is \( \mathcal{O}(N^2 d_c) \). However, in practice, since the data are processed in batches, \( N \) typically represents the number of molecular maps in a fixed-size batch, denoted by \( b \), rather than the total number of molecules. In this case, \( b \) is much smaller than \( N \), and \( \log b < d_c \), which allows the time complexity of hypergraph construction to be rewritten as \( \mathcal{O}(b^2 d_c) \). The cost of constructing the hypergraph is negligible compared to the overall computational complexity of the model.
\end{remark}

\begin{table*}[ht]
\renewcommand{\arraystretch}{1}
  \centering
  \caption{ROC-AUC (\%, $\Uparrow$) performance of various baseline methods on eight molecular attribute prediction classification datasets. ``$\Uparrow$" indicates that higher ROC-AUC values correspond to superior model performance. The results are presented as the mean $\pm$ standard deviation of ROC-AUC scores across five random seeds, with values ranging from 0 to 4. The best and second-best results are highlighted in \textbf{bold} and \underline{underlined}, respectively. ``$\uparrow$" denotes performance improvement. ``$\Delta$" denotes the absolute percentage improvement, calculated as \(\Delta = \text{ROC-AUC}_{\text{MMSA}} - \text{ROC-AUC}_{\text{baseline}}\).}
  \label{tab1}
  \begin{adjustbox}{max width=\textwidth}
\begin{tabular}{ccccccccccc}
\toprule
            & Tox21             & ToxCast           & Sider             & ClinTox           & MUV               & HIV               & BBBP              & BACE              & Average     & $\Delta$          \\ \midrule
\#Molecules & 7831              & 8575              & 1427              & 1478              & 93087             & 41127             & 2039              & 1513              & -           & -                 \\ \midrule
No pre-train & 72.9 $\pm$ 0.7          & 59.4 $\pm$ 0.4          & 58.6 $\pm$ 0.6          & 56.8 $\pm$ 1.7          & 70.3 $\pm$ 0.5          & 71.6 $\pm$ 0.8          & 65.7 $\pm$ 0.7          & 72.1 $\pm$ 0.2          & 65.93       & $\uparrow$10.01\% \\ \midrule
GIN         & 73.2 $\pm$ 0.6          & 59.8 $\pm$ 0.5          & 58.7 $\pm$ 1.0          & 57.4 $\pm$ 1.1          & 71.8 $\pm$ 1.2          & 74.2 $\pm$ 0.4          & 66.0 $\pm$ 0.5          & 69.6 $\pm$ 0.2          & 66.34       & $\uparrow$9.60\%  \\
EdgePred    & 76.0 $\pm$ 0.6          & 64.1 $\pm$ 0.6          & 60.4 $\pm$ 0.7          & 64.1 $\pm$ 3.7          & 75.1 $\pm$ 1.2          & 76.3 $\pm$ 1.0          & 67.3 $\pm$ 2.4          & 77.3 $\pm$ 3.5          & 70.08       & $\uparrow$5.86\%  \\
AttrMask & 73.6 $\pm$ 0.3          & 62.6 $\pm$ 0.6          & 59.7 $\pm$ 1.8          & 74.0 $\pm$ 3.4          & 72.5 $\pm$ 1.5          & 75.6 $\pm$ 1.0          & 70.6 $\pm$ 1.5          & 78.8 $\pm$ 1.2          & 70.93       & $\uparrow$5.01\%  \\
GPT-GNN     & 74.9 $\pm$ 0.3          & 62.5 $\pm$ 0.4          & 58.1 $\pm$ 0.3          & 58.3 $\pm$ 5.2          & 75.9 $\pm$ 2.3          & 75.2 $\pm$ 2.1          & 64.5 $\pm$ 1.4          & 77.9 $\pm$ 3.2          & 68.41       & $\uparrow$7.53\%  \\
InfoGraph   & 73.3 $\pm$ 0.6          & 61.8 $\pm$ 0.4          & 58.7 $\pm$ 0.6          & 75.4 $\pm$ 4.3          & 74.4 $\pm$ 1.8          & 74.2 $\pm$ 0.9          & 68.7 $\pm$ 0.6          & 74.3 $\pm$ 2.6          & 70.10        & $\uparrow$5.84\%  \\
GraphLoG    & 75.0 $\pm$ 0.6          & 63.4 $\pm$ 0.6          & 59.6 $\pm$ 1.9          & 75.7 $\pm$ 2.4          & 75.5 $\pm$ 1.6          & 76.1 $\pm$ 0.8          & 68.7 $\pm$ 1.6          & 78.6 $\pm$ 1.0          & 71.58       & $\uparrow$4.36\%  \\
G-Motif     & 73.6 $\pm$ 0.7          & 62.3 $\pm$ 0.6          & 61.0 $\pm$ 1.5          & 77.7 $\pm$ 2.7          & 73.0 $\pm$ 1.8          & 73.8 $\pm$ 1.2          & 66.9 $\pm$ 3.1          & 73.0 $\pm$ 3.3          & 70.16       & $\uparrow$5.78\%  \\
JOAO        & 74.8 $\pm$ 0.6          & 62.8 $\pm$ 0.7          & 60.4 $\pm$ 1.5          & 66.6 $\pm$ 3.1          & 76.6 $\pm$ 1.7          & 76.9 $\pm$ 0.7          & 66.4 $\pm$ 1.0          & 73.2 $\pm$ 1.6          & 69.71       & $\uparrow$6.23\%  \\
GraphMVP  & 74.4 $\pm$ 0.2          & 63.1 $\pm$ 0.2          & 62.2 $\pm$ 1.1          & 77.8 $\pm$ 2.8          & 75.4 $\pm$ 0.6          & 76.5 $\pm$ 0.1          & 71.4 $\pm$ 0.5          & 79.3 $\pm$ 1.5          & 72.51       & $\uparrow$3.43\%  \\
GraphMAE    & 75.2 $\pm$ 0.9          & 63.6 $\pm$ 0.3          & 60.5 $\pm$ 1.2          & 76.5 $\pm$ 3.0          & 76.4 $\pm$ 2.0          & 76.8 $\pm$ 0.6          & 71.2 $\pm$ 1.0          & 78.2 $\pm$ 1.5          & 72.30        & $\uparrow$3.64\%  \\
MoleculeSDE & 76.2 $\pm$ 0.3          & 65.0 $\pm$ 0.2          & 60.8 $\pm$ 0.3          & \textbf{83.2 $\pm$ 0.5} & \underline{78.9 $\pm$ 0.3}    & 78.5 $\pm$ 0.9          & \underline{71.5 $\pm$ 0.7}    & 79.1 $\pm$ 2.1          & \underline{74.15} & $\uparrow$1.80\%  \\
Mole-BERT   & \underline{76.4 $\pm$ 0.4}    & 64.1 $\pm$ 0.2          & 61.6 $\pm$ 0.8          & 77.6 $\pm$ 2.5          & 77.2 $\pm$ 1.4          & 77.0 $\pm$ 0.5          & 71.3 $\pm$ 1.3          & 79.8 $\pm$ 1.4          & 73.13       & $\uparrow$2.81\%  \\
IME         & 75.2 $\pm$ 0.4          & \underline{65.4 $\pm$ 0.3}    & \underline{62.5 $\pm$ 0.8}    & 72.2 $\pm$ 1.4          & 78.4 $\pm$ 1.5          & \underline{78.8 $\pm$ 0.6}    & 68.1 $\pm$ 1.0          & \underline{81.6 $\pm$ 0.9}    & 72.78       & $\uparrow$3.16\%  \\
\textbf{MMSA (Ours)}        & \textbf{77.6 $\pm$ 0.5} & \textbf{66.2 $\pm$ 0.6} & \textbf{64.3 $\pm$ 0.5} & \underline{82.6 $\pm$ 1.2}    & \textbf{80.3 $\pm$ 1.0} & \textbf{79.6 $\pm$ 0.8} & \textbf{74.5 $\pm$ 0.7} & \textbf{82.4 $\pm$ 0.4} & \textbf{75.94}       & -                 \\ \bottomrule
\end{tabular}
\end{adjustbox}
\vspace{-0.5em}
\end{table*}

\section{Experiments}
\label{Experiment}
\subsection{Experimental Setting}
\subsubsection{Datasets}
We pre-train the model on the same dataset and fine-tune it on various downstream tasks. We randomly selected 100K eligible molecules with 2D and 3D structures from the Drugs dataset \cite{axelrod2022geom} for pre-training. Since molecular conformations better reflect the properties of molecules, we follow previous work by considering five conformations for each molecule. For the downstream tasks, the primary task is molecular property prediction. We use eight widely used binary classification datasets in MoleculeNet \cite{wu2018moleculenet} and evaluate the model using the ROC-AUC metric. To conduct a more comprehensive evaluation, we test the effectiveness of MMSA on four regression datasets from GraphMVP \cite{liu2021pre}, using the root mean square error (RMSE) metric. Additionally, we perform molecular retrieval both within the same dataset and across datasets to test whether MMSA effectively learns a broad range of molecular structures. We employ scaffold splitting \cite{ramsundar2019deep} to partition molecules based on their structural scaffolds, simulating real-world usage scenarios. 

\subsubsection{Baselines}
Our baselines cover three typical categories in molecular pre-training. The first category consists of methods based on 2D graph representations, such as GIN \cite{xu2018powerful}, EdgePred \cite{hamilton2017inductive}, AttrMask \cite{hu2020strategies}, GPT-GNN \cite{hu2020gpt}, InfoGraph \cite{sun2019infograph}, GraphLoG \cite{xu2021self}, G-Motif \cite{rong2020self}, JOAO \cite{you2021graph}, GraphMAE \cite{hou2022graphmae}, and Mole-BERT \cite{xiamole}. The second category includes multi-modal pre-training methods designed to integrate both 2D graphs and 3D conformations of molecules, such as GraphMVP \cite{liu2021pre} and MoleculeSDE \cite{liu2023group}. The third category comprises multi-modal pre-training methods incorporating molecular images, such as IME \cite{xiangimage}.

\subsubsection{Implementation details} 
In the pre-training phase, we use a five-layer GIN to extract the 2D graph representations of molecules, a pre-trained ResNet-18 to extract the 2D image representations of molecules, and a four-layer ComENet to extract the 3D graph representations. Considering the missing connections between molecules, we employ the widely used $K$-nearest neighbor (KNN) algorithm to estimate higher-order correlations, setting \( K = 10 \). In the experiments, the memory bank contains 128 memory anchor points, with each anchor point having a dimensionality of \( \bm{a}_j \in \mathbb{R}^{256} \), and all molecular embeddings are 256-dimensional vectors. The model is pre-trained for 30 epochs with a batch size of 128 and a learning rate 0.001. The hyperparameters \( \lambda \) and \( \alpha \) are set to 0.6 and 0.5, respectively. For fair comparison and consistent experimental setup, we use the five-layer GIN model with 256 hidden dimensions as the baseline model for downstream task evaluation. Following the setup of \cite{hu2020gpt}, we train for 100 epochs with a learning rate of 0.001 and report the average performance and standard deviation over five runs. Notice that the results of some baselines may differ from their original papers because of inconsistent evaluation settings, and we reproduced them with the same evaluation to ensure comparability.

\begin{figure*}[t]
  \begin{minipage}[c]{0.68\textwidth}
  \centering
  \renewcommand{\arraystretch}{1.15}
  \captionof{table}{RMSE ($\Downarrow$) performance of various methods on four regression datasets for molecular property prediction. ``$\Downarrow$" indicates that lower RMSE values correspond to better model performance. The results are reported as the mean (standard deviation) of RMSE across five random seeds. 
  ``$\Delta$" represents the absolute percentage improvement, calculated as \(\Delta = \left( 1 - \text{RMSE}_{\text{baseline
  }}/\text{RMSE}_{\text{MMSA}} \right) \times 100\).}
    \label{tab2}
\begin{adjustbox}{max width=0.95\textwidth}
\begin{tabular}{ccccccc}
\toprule
            & ESOL                 & Lipo                 & Malaria              & CEP                  & Average           & $\Delta$          \\ \hline
\#Molecules & 1128                 & 4200                 & 9999                 & 29978                & -                 & -                 \\ \hline
No pre-train & 1.178 (0.044)          & 0.797 (0.004)          & 1.135 (0.003)          & 1.296 (0.036)          & 1.102             & $\uparrow$9.65\%  \\ \hline
GIN         & 1.379 (0.034)          & 0.821 (0.027)          & 1.115 (0.012)          & 1.342 (0.015)          & 1.164             & $\uparrow$15.82\% \\
EdgePred    & 1.295 (0.021)          & 0.752 (0.015)          & 1.124 (0.015)          & 1.305 (0.008)          & 1.110              & $\uparrow$10.45\% \\
AttrMask    & 1.318 (0.027)          & 0.787  (0.012)          & 1.119 (0.015)          & 1.326 (0.008)          & 1.138             & $\uparrow$13.23\% \\
JOAO        & 1.165 (0.015)          & 0.728 (0.021)          & 1.147 (0.015)          & 1.293 (0.016)          & 1.083             & $\uparrow$7.76\%  \\
GraphMVP    & 1.114 (0.020)          & 0.719  (0.012)          & 1.114 (0.015)          & 1.236 (0.021)          & 1.042             & $\uparrow$3.68\%  \\
GraphMAE    & 1.116 (0.006)          & 0.729 (0.018)          & 1.123 (0.014)          & 1.262 (0.008)          & 1.058             & $\uparrow$5.27\%  \\
MoleculeSDE & \underline{1.058 (0.024)}    & 0.712 (0.015)          & \underline{1.084 (0.013)}    & \underline{1.195 (0.011)}    & \underline{1.012}       & $\uparrow$0.70\%  \\
Mole-BERT   & 1.090 (0.035)          & \underline{0.710 (0.006)}    & 1.093 (0.009)          & 1.232 (0.009)          & 1.018             & $\uparrow$1.29\%  \\
IME         & 1.069 (0.032)          & 0.714 (0.012)          & 1.091 (0.012)          & 1.288 (0.014)          & 1.041             & $\uparrow$3.58\%  \\
\textbf{MMSA (Ours)}        & \textbf{1.042 (0.017)} & \textbf{0.704 (0.003)} & \textbf{1.078 (0.005)} & \textbf{1.187 (0.006)} & \textbf{1.005}    & -    \\ \bottomrule
\end{tabular}
\end{adjustbox}
\end{minipage}
\hfill
  \begin{minipage}[c]{0.3\linewidth}
    \centering
\includegraphics[width=0.98\textwidth]{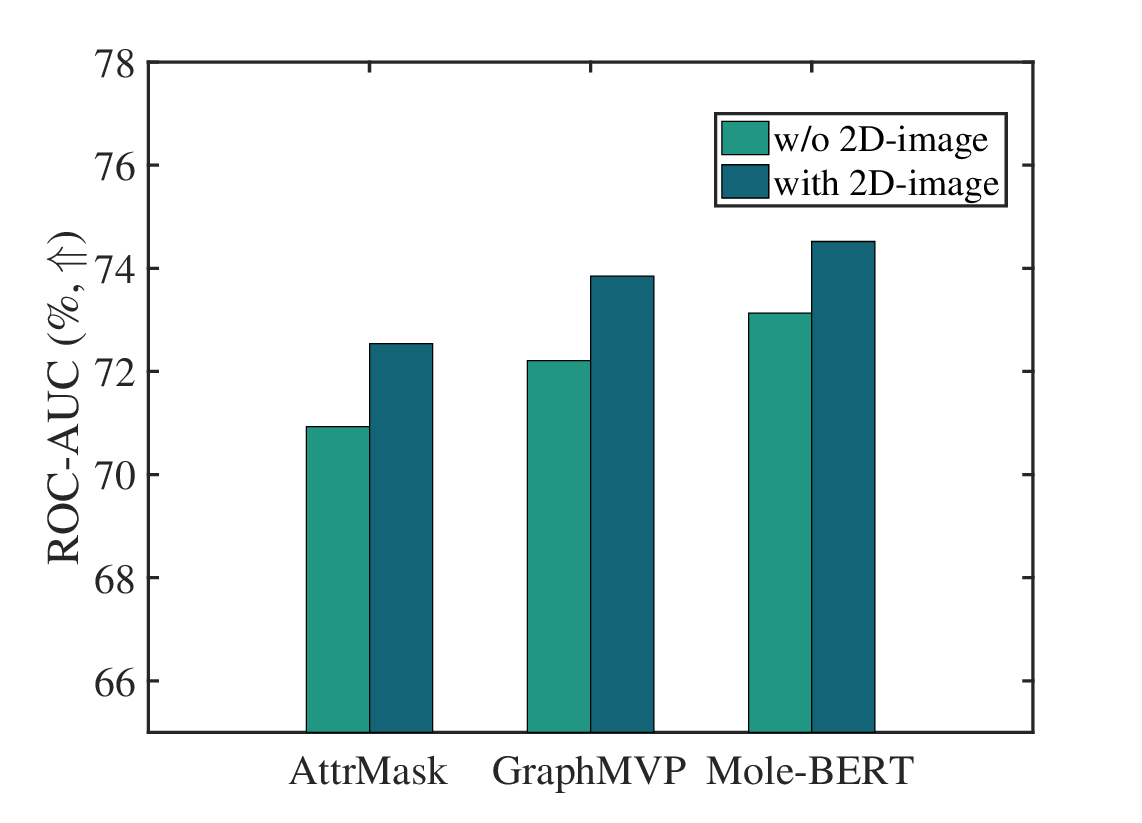}
\caption{Average ROC-AUC performance (\%, $\Uparrow$) on eight classification-based attribute prediction tasks, with and without using 2D image data. The addition of 2D image data results in a significant performance improvement.}
\label{fig3}
\end{minipage}
\vspace{-0.6em}
\end{figure*}

\begin{table*}[!t]
\renewcommand{\arraystretch}{1.0}
\centering
\caption{Molecular retrieval task in the same dataset. The query molecule and the four closest molecules to the extracted representation from the same dataset. The similarity values are displayed below the corresponding visualized molecules.}
\label{tab3}
\begin{adjustbox}{max width=1\textwidth}
\begin{tabular}{|c|c|c|c|c|c|}
\hline
\multirow{3}{*}{\begin{tabular}[c]{@{}c@{}}Query Molecular\\ \includegraphics[width=0.2\textwidth]{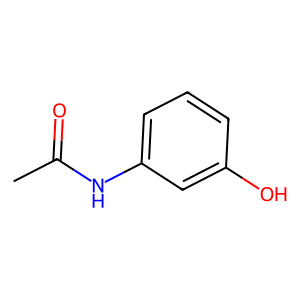} \\ CC(=O)Nc1cccc(O)c1\end{tabular}} 
& \textbf{ECFP}    & \begin{tabular}[c]{@{}c@{}} \\ \includegraphics[width=0.16\textwidth]{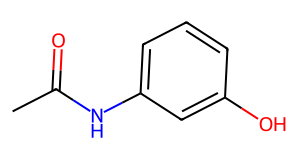} \\ 1.0000 \\ \end{tabular} 
& \begin{tabular}[c]{@{}c@{}} \\ \includegraphics[width=0.16\textwidth]{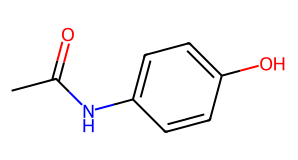} \\ 0.7715 \\ \end{tabular} 
& \begin{tabular}[c]{@{}c@{}} \\ \includegraphics[width=0.16\textwidth]{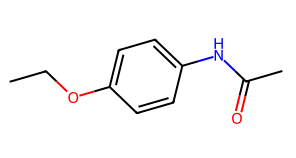} \\ 0.6236  
  \end{tabular} 
& \begin{tabular}[c]{@{}c@{}} \\ \includegraphics[width=0.16\textwidth]{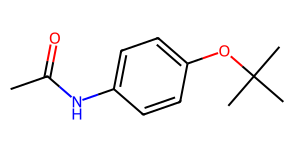} \\ 0.6110 \\ \end{tabular} \\ \cline{2-6} 
& \textbf{No pre-train} 
& \begin{tabular}[c]{@{}c@{}} \\ \includegraphics[width=0.16\textwidth]{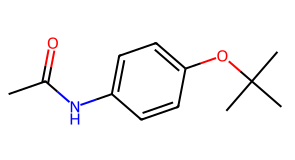} \\ 0.9985 \\ \end{tabular}  
& \begin{tabular}[c]{@{}c@{}} \\ \includegraphics[width=0.16\textwidth]{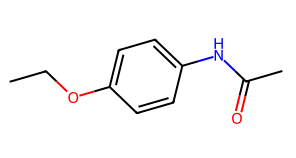} \\ 0.9821 \\ \end{tabular}  
& \begin{tabular}[c]{@{}c@{}} \\ \includegraphics[width=0.16\textwidth]{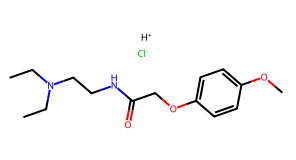} \\ 0.9732 \\ \end{tabular}  
& \begin{tabular}[c]{@{}c@{}} \\ \includegraphics[width=0.16\textwidth]{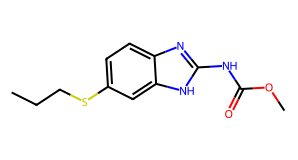} \\ 0.9471 \\ \end{tabular} \\ \cline{2-6} 
& \textbf{MMSA (Ours)}       
& \begin{tabular}[c]{@{}c@{}} \\ \includegraphics[width=0.16\textwidth]{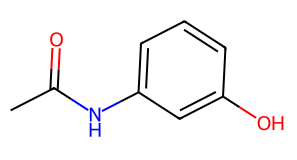} \\ 0.9874  \\
 \end{tabular}  
& \begin{tabular}[c]{@{}c@{}} \\ \includegraphics[width=0.16\textwidth]{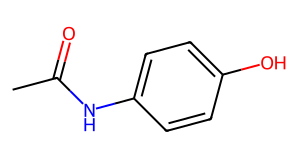} \\ 0.9727 \\ \end{tabular} 
& \begin{tabular}[c]{@{}c@{}} \\ \includegraphics[width=0.16\textwidth]{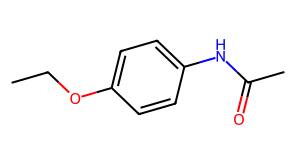} \\ 0.8751 \\ \end{tabular}  
& \begin{tabular}[c]{@{}c@{}} \\ \includegraphics[width=0.16\textwidth]{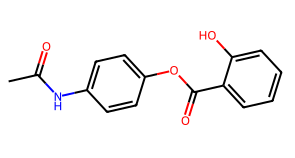} \\ 0.8107 \\ \end{tabular} \\ \hline
\end{tabular}
\end{adjustbox}
\end{table*}

\subsection{Main Results}
Table \ref{tab1} presents the performance of our proposed method across eight biological classification tasks from the MoleculeNet benchmark. Notably, our approach achieves state-of-the-art results in seven of the eight tasks. On average, our method outperforms the second-best model, MoleculeSDE, by a significant margin, with an average ROC-AUC score of 75.94 compared to 74.15 for MoleculeSDE. This improvement underscores the effectiveness of our method in learning more comprehensive and discriminative molecular representations, particularly for tasks involving complex biological data.

Additionally, we expand the evaluation to include broader drug discovery tasks and precisely four regression benchmarks, as shown in Table \ref{tab2}. The results follow the same trend of superior performance, with our method consistently outperforming all other models and achieving the best results across all regression tasks. It further confirms that the learned representations—capturing both molecular structure and semantic information—are highly advantageous for classification tasks and regression-based predictions, which often require precise modeling of molecular properties. In summary, the consistent superiority of MMSA can be attributed to integrating a structure-awareness module and memory mechanisms. These innovations allow our model to capture higher-order interactions and intrinsic molecular properties, essential for accurately predicting biological and chemical properties.

\begin{table*}[!t]
\renewcommand{\arraystretch}{1.0}
\centering
\caption{Molecular retrieval task across datasets. The query molecule and the four closest molecules to the extracted representation from different datasets. The similarity values are displayed below the corresponding visualized molecules.}
\label{tab4}
\begin{adjustbox}{max width=1\textwidth}
\begin{tabular}{|c|c|c|c|c|c|}
\hline
\multirow{3}{*}{\begin{tabular}[c]{@{}c@{}} \\ \\ \\ \\Query Molecular\\ \includegraphics[width=0.2\textwidth]{pic/query_molecule.png} \\ CC(=O)Nc1cccc(O)c1\end{tabular}} 
& \textbf{ECFP}    
& \begin{tabular}[c]{@{}c@{}}  \\ \includegraphics[width=0.16\textwidth]{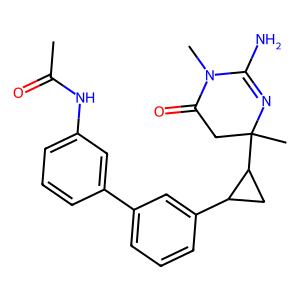} \\ 0.9998 \\ \end{tabular} 
& \begin{tabular}[c]{@{}c@{}}  \\ \includegraphics[width=0.16\textwidth]{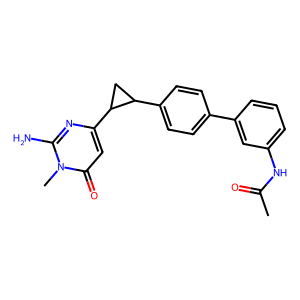} \\ 0.9264 \\ \end{tabular} 
& \begin{tabular}[c]{@{}c@{}}   \\ \includegraphics[width=0.16\textwidth]{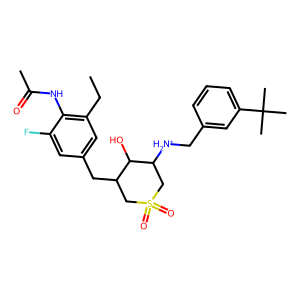} \\ 0.8058  
  \end{tabular} 
& \begin{tabular}[c]{@{}c@{}}  \includegraphics[width=0.16\textwidth]{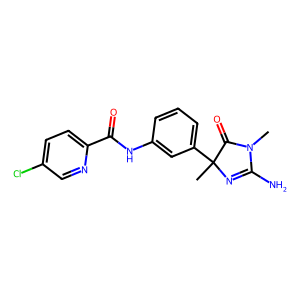} \\ 0.7885 \\ \end{tabular} \\ \cline{2-6} 
& \textbf{No pre-train} 
& \begin{tabular}[c]{@{}c@{}}  \\ \includegraphics[width=0.16\textwidth]{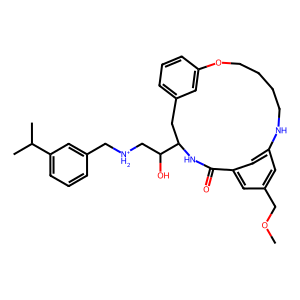} \\ 0.9967 \\ \end{tabular}  
& \begin{tabular}[c]{@{}c@{}}  \\ \includegraphics[width=0.16\textwidth]{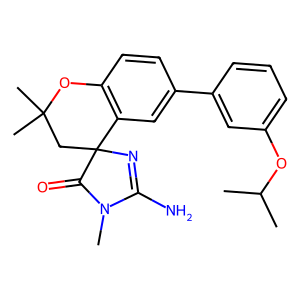} \\ 0.9915 \\ \end{tabular}  
& \begin{tabular}[c]{@{}c@{}}  \\ \includegraphics[width=0.16\textwidth]{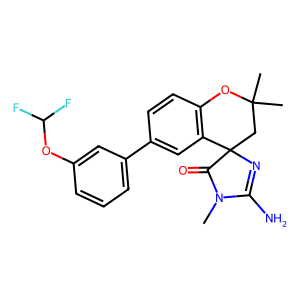} \\ 0.9824 \\ \end{tabular}  
& \begin{tabular}[c]{@{}c@{}}  \\ \includegraphics[width=0.16\textwidth]{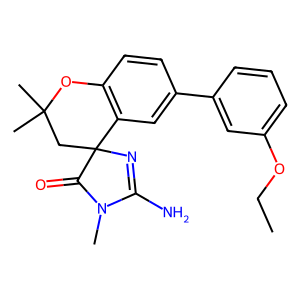} \\ 0.9761 \\ \end{tabular} \\ \cline{2-6} 
& \textbf{MMSA (Ours)}       
& \begin{tabular}[c]{@{}c@{}}  \\ \includegraphics[width=0.16\textwidth]{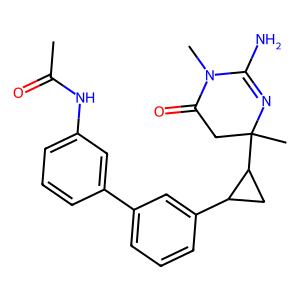} \\ 0.9996  \\
 \end{tabular}  
& \begin{tabular}[c]{@{}c@{}} \\ \includegraphics[width=0.16\textwidth]{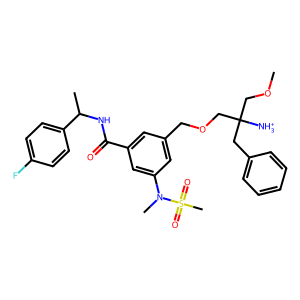} \\ 0.9869 \\ \end{tabular} 
& \begin{tabular}[c]{@{}c@{}}  \\ \includegraphics[width=0.16\textwidth]{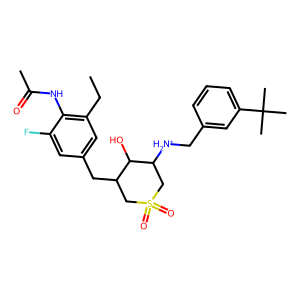} \\ 0.9703 \\ \end{tabular}  
& \begin{tabular}[c]{@{}c@{}}  \\ \includegraphics[width=0.16\textwidth]{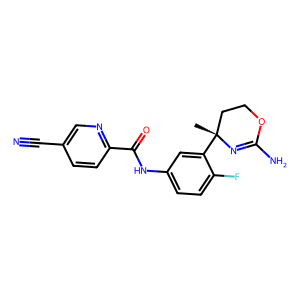} \\ 0.9657 \\ \end{tabular} \\ \hline
\end{tabular}
\end{adjustbox}
\end{table*}

\subsection{Molecule Retrieval}
To conduct a more comprehensive evaluation, we extract the molecular representation of a query molecule from BBBP and compute the similarity between the query molecule and all reference molecules in the dataset. Table \ref{tab3} presents the four molecules most similar to the query molecule and their corresponding similarities. Notably, the molecular representations generated by MMSA exhibit a strong alignment with fingerprint similarities (ECFP\footnote{ECFP (Extended Connectivity Fingerprint) is a molecular fingerprinting algorithm that recursively generates topological features of a molecule, capturing the local structural environment, and is commonly used for chemical similarity analysis and machine learning tasks.}), showing that MMSA has learned meaningful and chemically relevant representations. It indicates that MMSA effectively captures critical structures and features of molecules, reinforcing its ability to generate discriminative embeddings for molecular similarity tasks.

To further assess the cross-dataset generalization ability of our method, we conduct a cross-dataset retrieval experiment. Specifically, we select a query molecule from the BBBP dataset and searched for structurally similar molecules in the BACE dataset. As shown in Table \ref{tab4}, the experimental results indicate that our method effectively identifies molecules with highly similar structures in the BACE dataset, despite differences in molecular distributions and task settings between the two datasets. In contrast, models without pre-training fail to retrieve structurally similar molecules in this cross-dataset retrieval task. This outcome highlights the strong generalization capability of the proposed method across different datasets. Our method MMSA not only excels in within-dataset tasks but also performs exceptionally well in cross-dataset retrieval scenarios. By effectively capturing the fundamental structural features and inherent patterns of molecules, our approach transcends the boundaries of specific datasets and learns molecular representations with broad adaptability. This enables effective transfer across diverse datasets, enhancing the adaptability and generalization capability of MMSA.

\subsection{Ablation Study}
We conduct an ablation study to validate the effectiveness of the proposed multi-modal molecular representation learning module and the structure-awareness module. 

\textbf{The multi-modal molecular representation module} aims to address the differences between modalities and generate a unified molecular representation. We perform the ablation analysis from two perspectives: one comparing the modal features and the other comparing the loss functions.
In the comparison of modal features, as shown in Table \ref{tab5}, the proposed method demonstrates significant advantages over using single-modal features and methods using 2D-3D multi-modal features. The features from different modalities show varying performance. Specifically, when the 2D graph modality is fixed, the 3D graph features outperform 2D graph features. This phenomenon can be attributed to the superior ability of 3D graphs to capture the spatial and stereochemical properties of molecules, leading to more accurate molecular representations. Additionally, to further validate the enhancement of molecular representations by the 2D graph modality, we incorporate 2D images into the baseline method based on 2D/3D graphs. As shown in Fig. \ref{fig3}, the performance of the model significantly improves when the image modality is included. In the comparison of loss functions, as shown in Table \ref{tab6}, the MMSA model achieves the best performance when both \(\mathcal{L}_{cl}\) and \(\mathcal{L}_{rl}\) loss functions are used in combination. This indicates that the proposed MMSA method can effectively integrate and utilize cross-modal collaborative information, thereby further improving the quality of molecular representations.

\begin{table}[!t]
\renewcommand{\arraystretch}{1.15}
  \centering
  \caption{Ablation Study. Performance (ROC-AUC \%, $\Uparrow$ and RMSE, $\Downarrow$) on MoleculeNet using data from different modals.}
  \label{tab5}
\begin{adjustbox}{max width=\linewidth}
\begin{tabular}{cccccccc}
\toprule
2D-G & 2D-I & 3D-G & BBBP ($\Uparrow$) & Tox21 ($\Uparrow$) & ClinTox ($\Uparrow$) & ESOL ($\Downarrow$)  & Lipo ($\Downarrow$) \\ \midrule
\checkmark       & -       & -       & 71.6 & 72.2  & 73.2    & 1.382 & 0.825 \\
\checkmark       & \checkmark       & -       & 72.8 & 76.1  & 75.1    & 1.122 & 0.732 \\
\checkmark       & -       & \checkmark       & 73.6 & 75.8  & 78.9    & 1.107 & 0.715 \\
\checkmark       & \checkmark       & \checkmark       & 74.5 & 77.3  & 80.6    & 1.042 & 0.704 \\
\bottomrule
\end{tabular}
\end{adjustbox}
\end{table}

\begin{table}[!t]
\renewcommand{\arraystretch}{1.15}
  \centering
  \caption{Ablation Study. Performance (ROC-AUC \%, $\Uparrow$ and RMSE, $\Downarrow$) on MoleculeNet using different components.}
  \label{tab6}
\begin{adjustbox}{max width=\linewidth}
\begin{tabular}{cccccc}
\toprule
                    & BBBP ($\Uparrow$) & Tox21 ($\Uparrow$) & ClinTox ($\Uparrow$) & Lipo ($\Downarrow$)  & ESOL ($\Downarrow$) \\ \midrule
w/o $\mathcal{L}_{cl}$    & 74.1 & 76.7  & 79.2    & 0.715 & 1.087 \\
w/o $\mathcal{L}_{rl}$    & 73.8 & 76.1  & 78.5    & 0.721 & 1.232 \\
w/o $\mathcal{L}_{me}$       & 73.1 & 74.9  & 78.1    & 0.713 & 1.185 \\
 w/o $\mathcal{L}_{pre}$& 74.0& 76.2& 78.4& 0.715&1.155\\ \midrule
GCN-based MMSA & 73.3 & 75.6  & 78.6    & 0.709 & 1.092 \\
MMSA                & 74.5 & 77.3  & 80.6    & 0.704 & 1.042 \\ \bottomrule
\end{tabular}
\end{adjustbox}
\vspace{-1em}
\end{table}

\begin{figure*}
    \centering
    \includegraphics[width=0.29\linewidth]{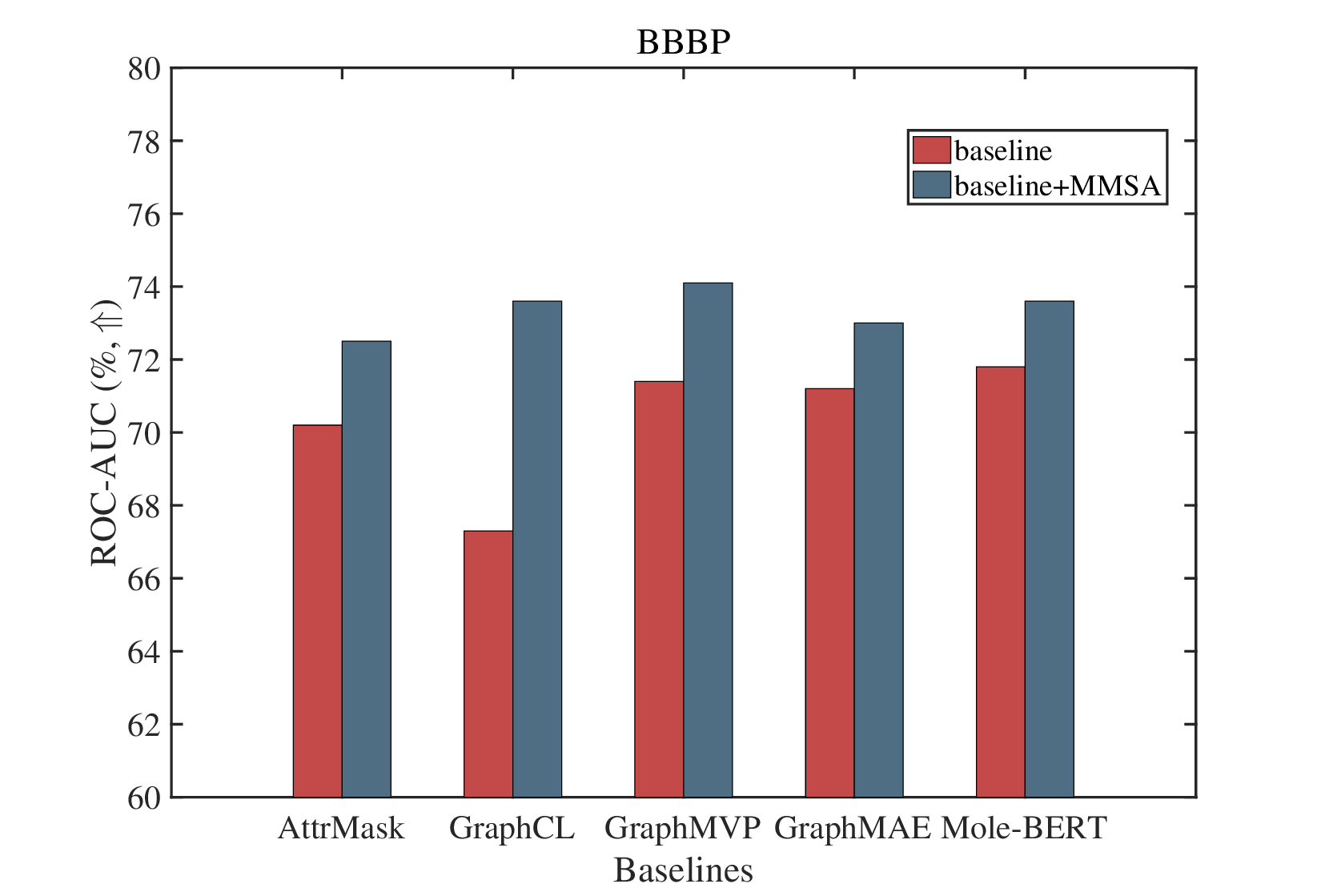}\hfill
    \includegraphics[width=0.29\linewidth]{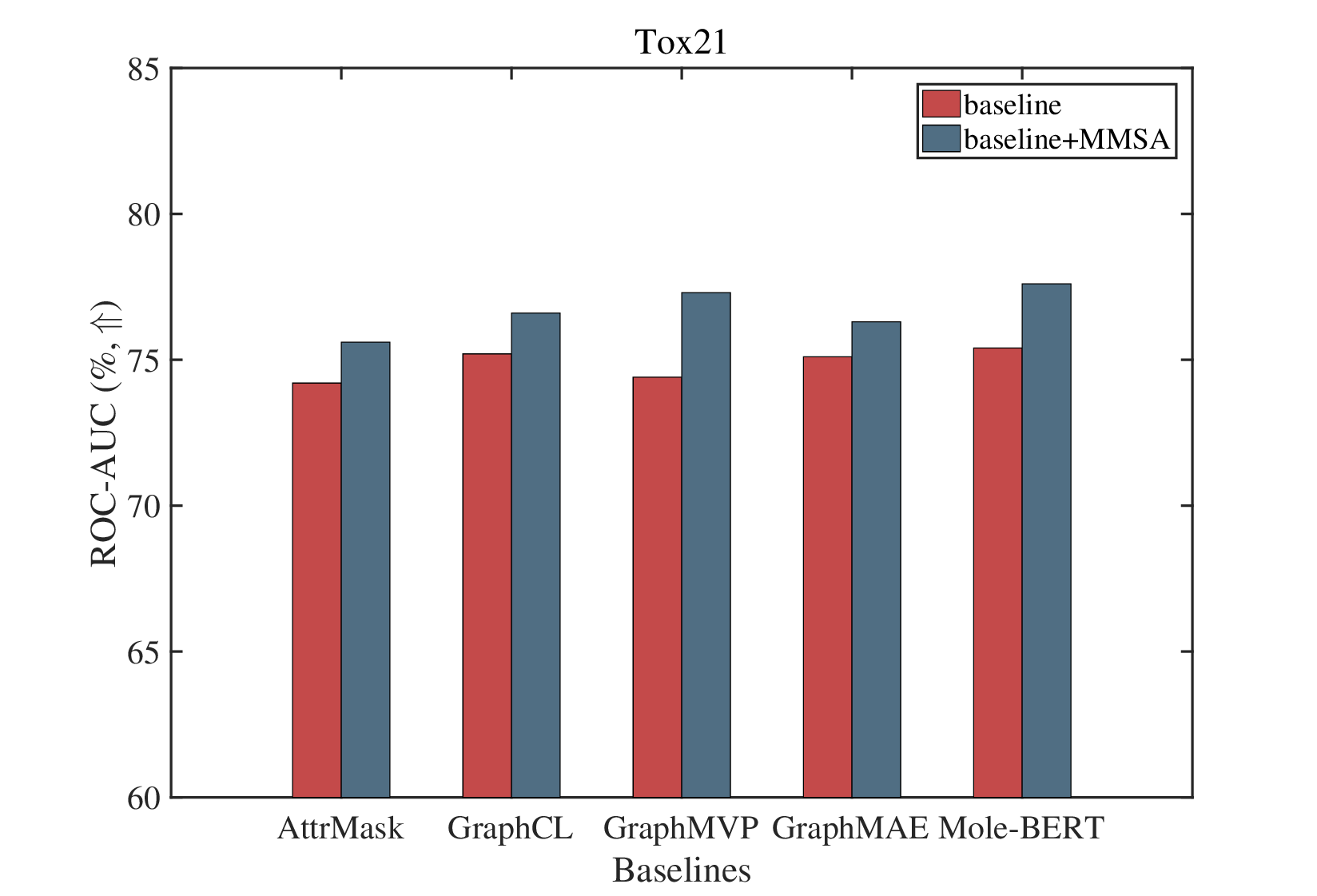}\hfill
    \includegraphics[width=0.29\linewidth]{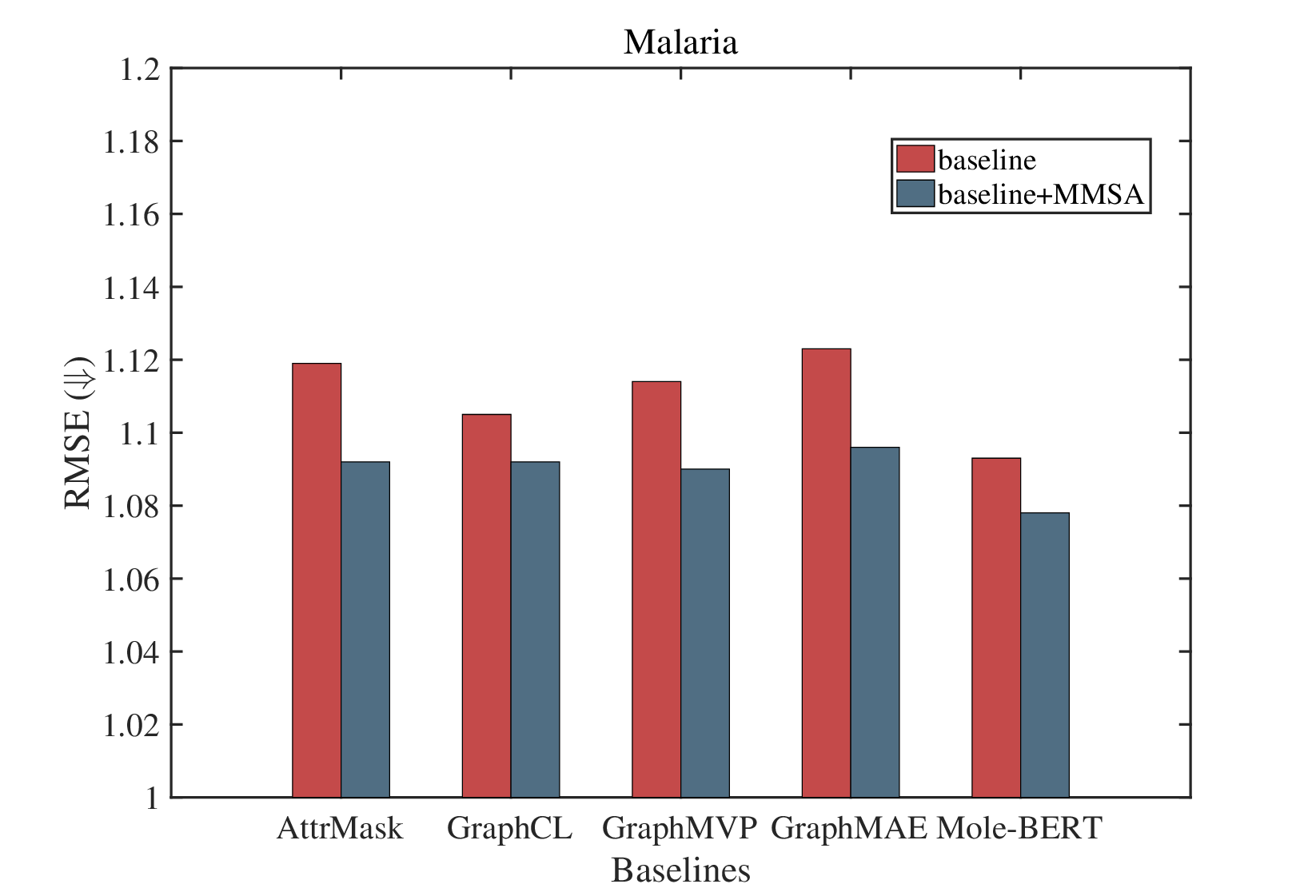}
    \caption{Performance (ROC-AUC \%, $\Uparrow$ and RMSE, $\Downarrow$) on MoleculeNet across different baseline methods equipped with MMSA.}
    \label{fig4}
\vspace{-1em}
\end{figure*}

\begin{figure*}[ht]
    \centering
    \includegraphics[width=0.24\linewidth]{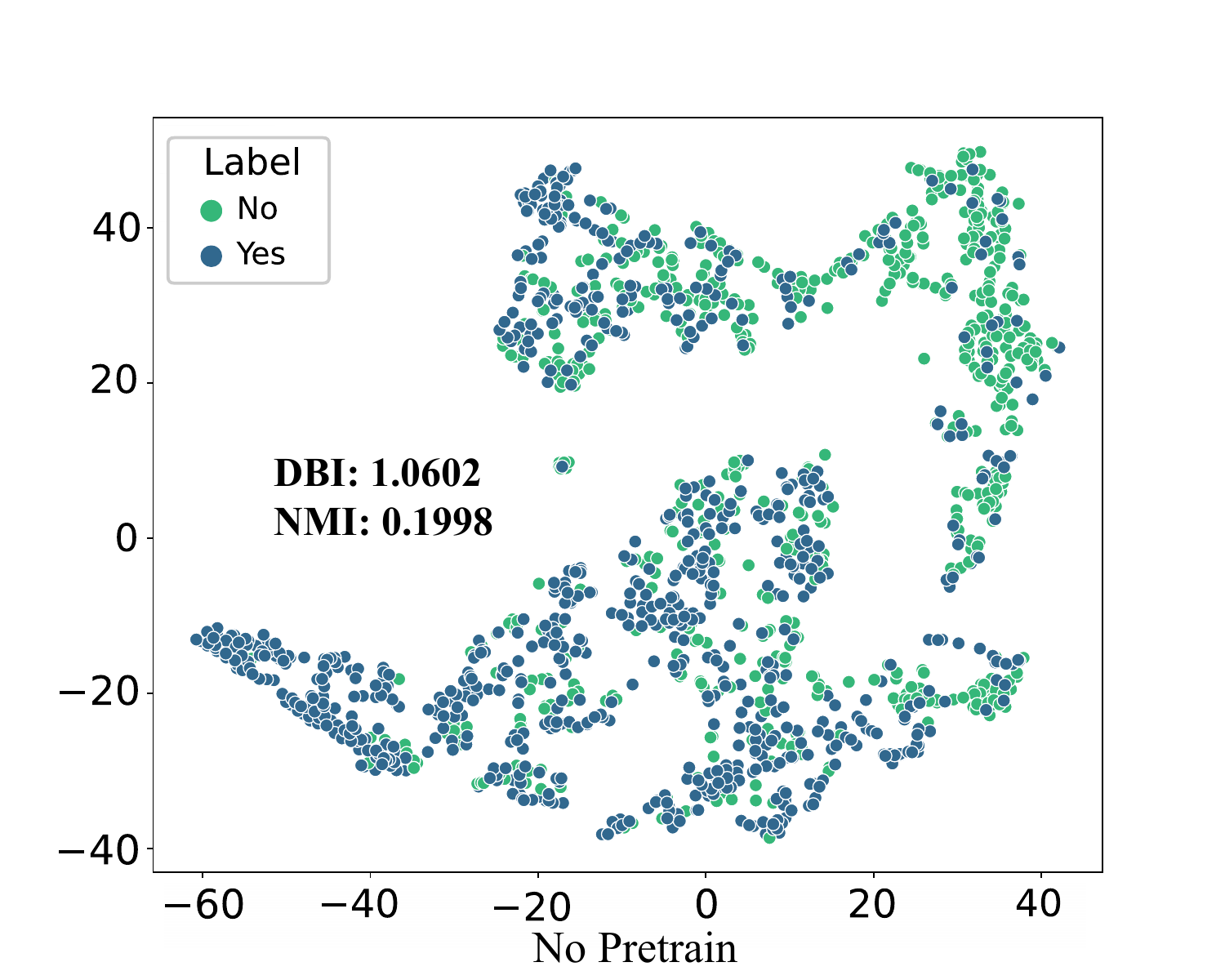}
    \includegraphics[width=0.24\linewidth]{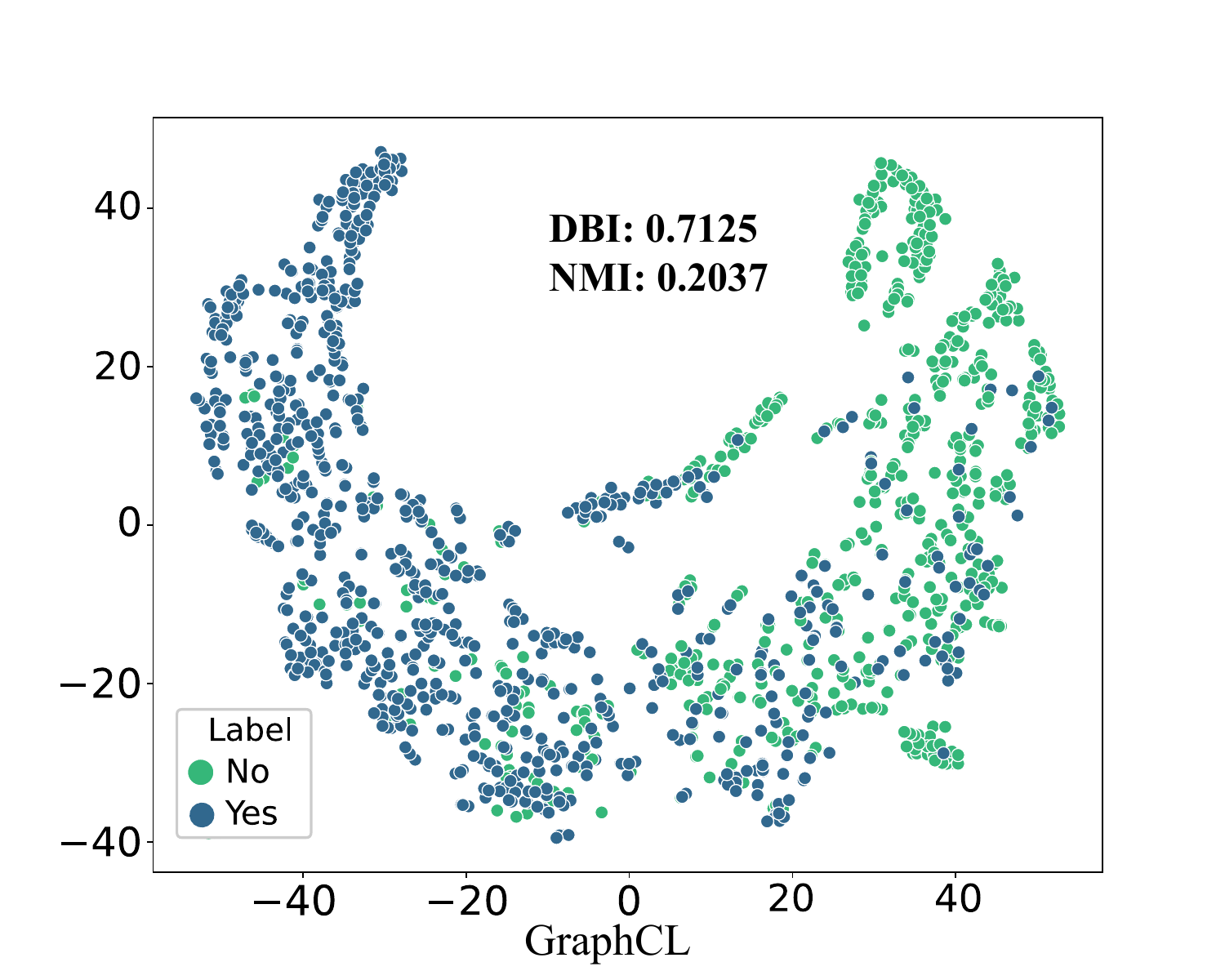}
    \includegraphics[width=0.24\linewidth]{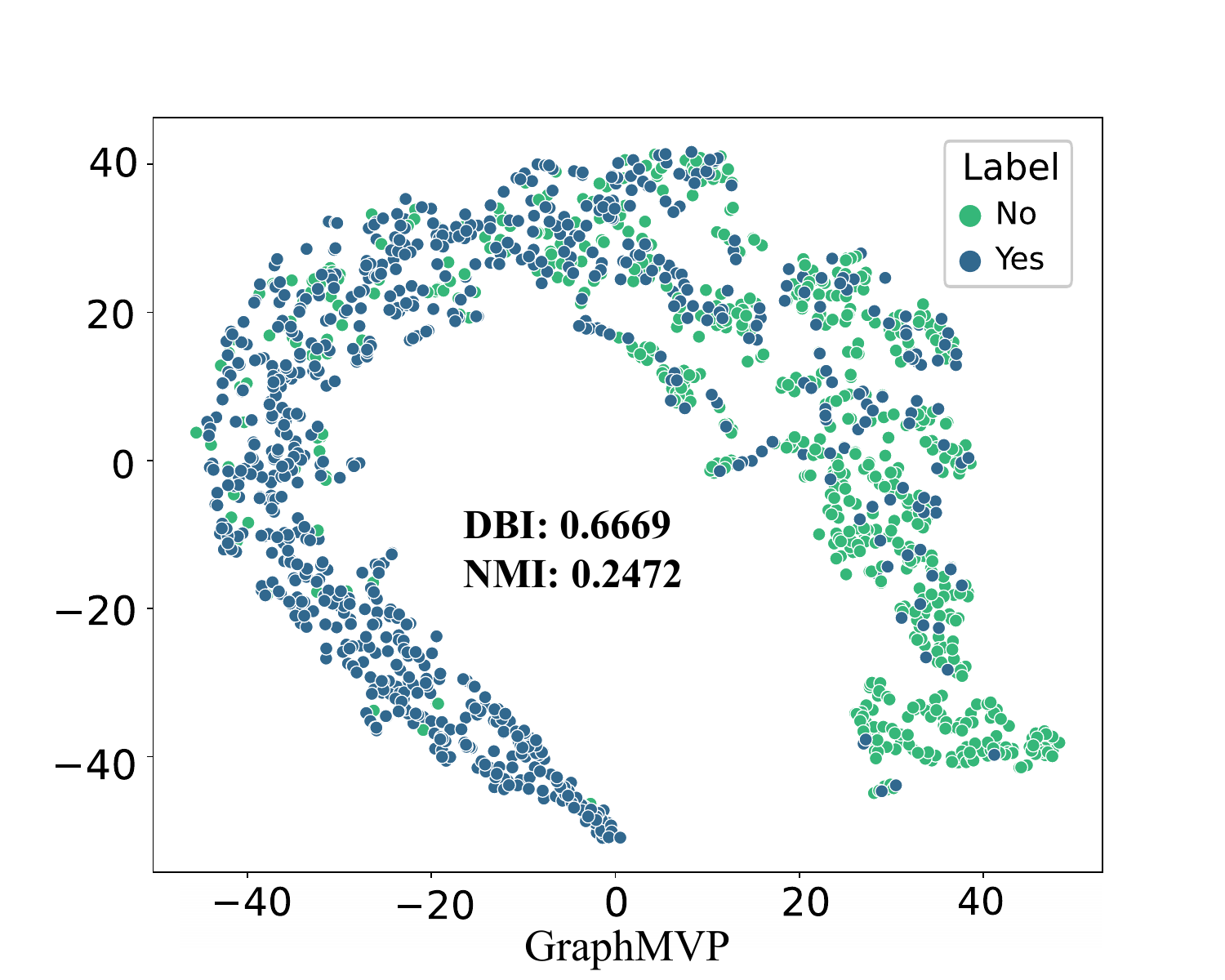}
    \includegraphics[width=0.24\linewidth]{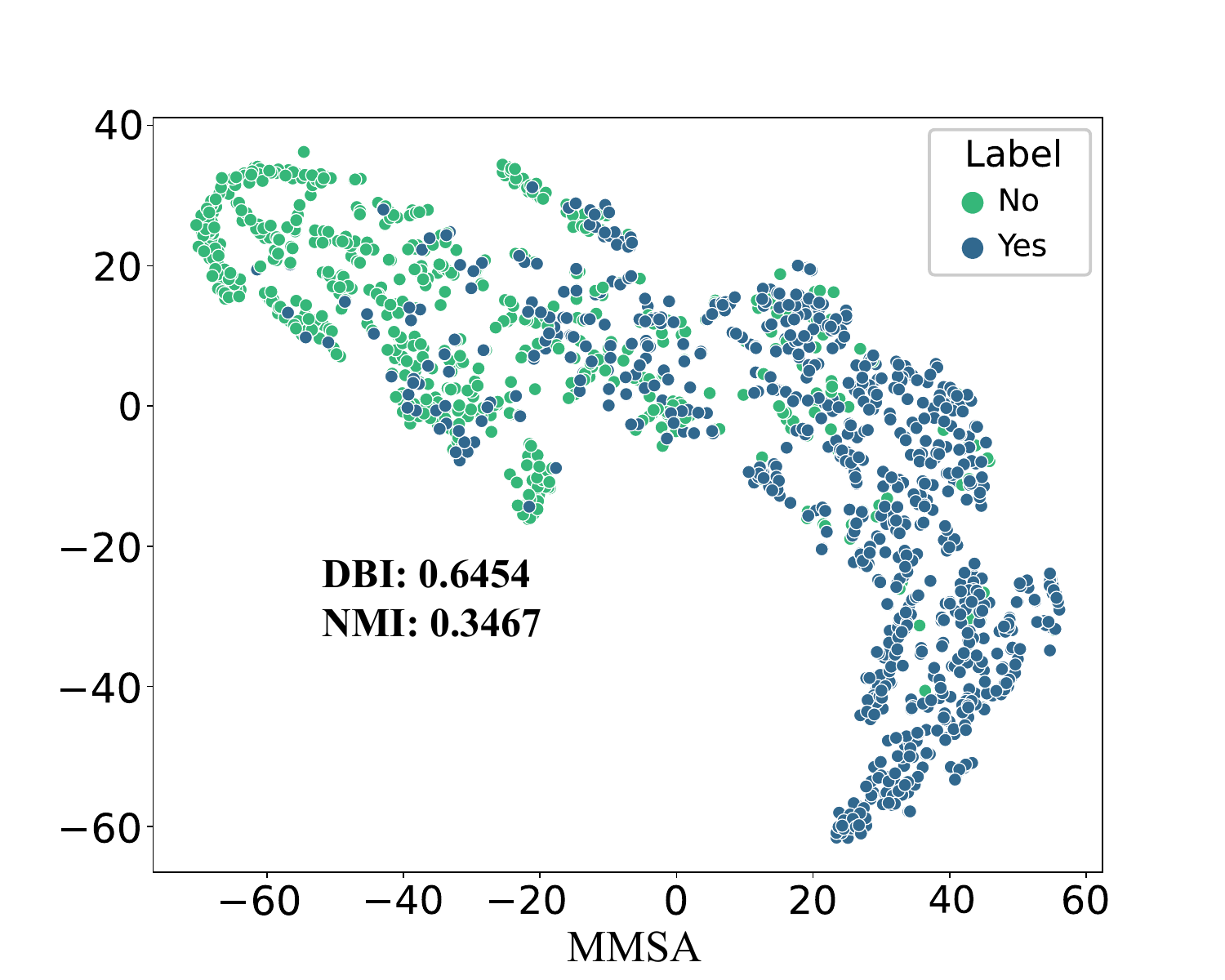}
    \caption{Clustering results on the BACE task using molecular representations obtained by different methods. The colours represent the labels of the downstream tasks (discrete binary labels for the BACE task). For each subgraph, the performance of the clustering results is evaluated using the Davies-Bouldin Index (DBI, where smaller values are better) and the Normalized Mutual Information (NMI, where larger values are better). The MMSA method (the last one) demonstrates a significant ability to improve clustering performance.}
    \label{fig5}
\vspace{-1.2em}
\end{figure*}

\textbf{In the structure-awareness module}, we perform an ablation study from two perspectives: the effectiveness of the hypergraph structure and the comparison of loss functions. We replace the hypergraph-based relational learning with a GCN. To ensure a fair comparison, the GCN model is configured to have the same number of layers and trainable parameters as the original hypergraph-based relational learning. Additionally, the graph construction for the GCN used the same $K$ value as the hypergraph-based relational learning with the $K$-nearest neighbor (KNN) algorithm. The experimental results in Table 6 indicate that the hypergraph-based MMSA performs best. Unlike graph edges, hyperedges in a hypergraph can model relationships beyond pairwise correlations, enabling the modeling of higher-order dependencies. This capability is crucial for capturing more complex structural patterns in molecular data. In the comparison of loss functions, the MMSA model with both \(\mathcal{L}_{me}\) and \(\mathcal{L}_{pre}\) losses achieves the best performance. Notably, MMSA without \(\mathcal{L}_{pre}\) outperforms the model without \(\mathcal{L}_{me}\), suggesting that the proposed method does not heavily depend on labeled data. This highlights the robustness of our approach, indicating that it can effectively learn from labeled and unlabeled data, further improving its generalization ability.

\begin{table}[t]
\renewcommand{\arraystretch}{1}
  \centering
  \caption{Comparison of pre-training performance (ROC-AUC \%, $\Uparrow$ and RMSE, $\Downarrow$) on MoleculeNet of different GNN models.}
  \label{tab7}
\begin{adjustbox}{max width=\linewidth}
\begin{tabular}{ccccc}
\toprule
            & BBBP ($\Uparrow$) & Tox21 ($\Uparrow$) & Lipo ($\Downarrow$)  & ESOL ($\Downarrow$) \\ \midrule
GCN         & 73.5 & 75.9  & 0.712 & 1.057 \\
GAT         & 73.9 & 76.8  & 0.715 & 1.051 \\
GIN         & 74.5 & 77.3  & 0.704 & 1.042 \\
GraphSage   & 73.5 & 76.5  & 0.710  & 1.054 \\
Graphformor & 74.7 & 77.0    & 0.713 & 1.047 \\ \bottomrule
\end{tabular}
\end{adjustbox}
\vspace{-1em}
\end{table}

\begin{figure}[t]
    \centering
    \includegraphics[width=0.42\linewidth]{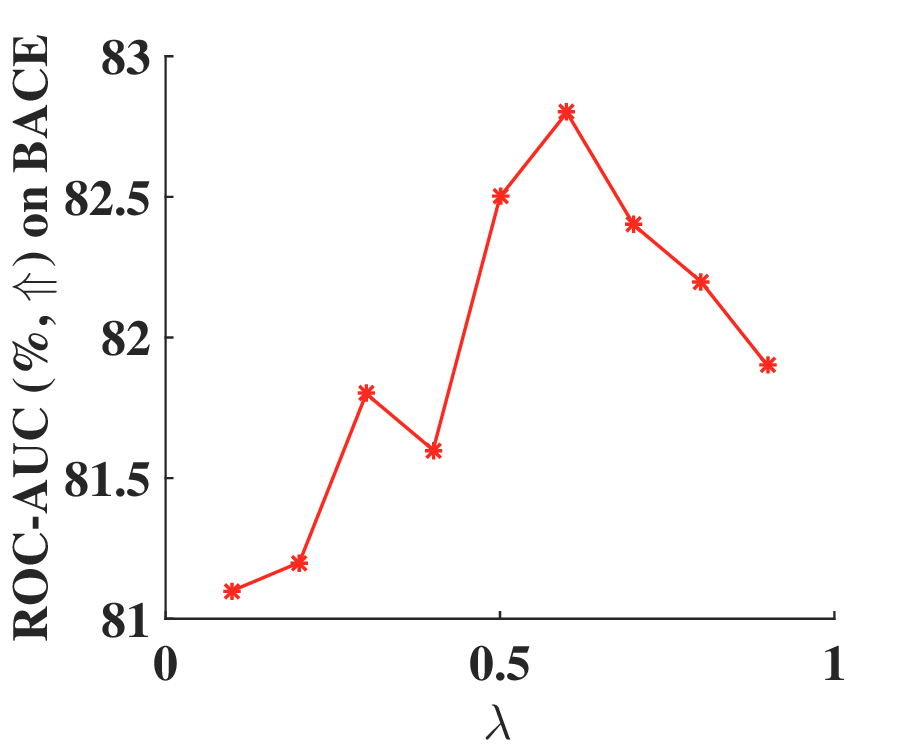}\hfill
    \includegraphics[width=0.42\linewidth]{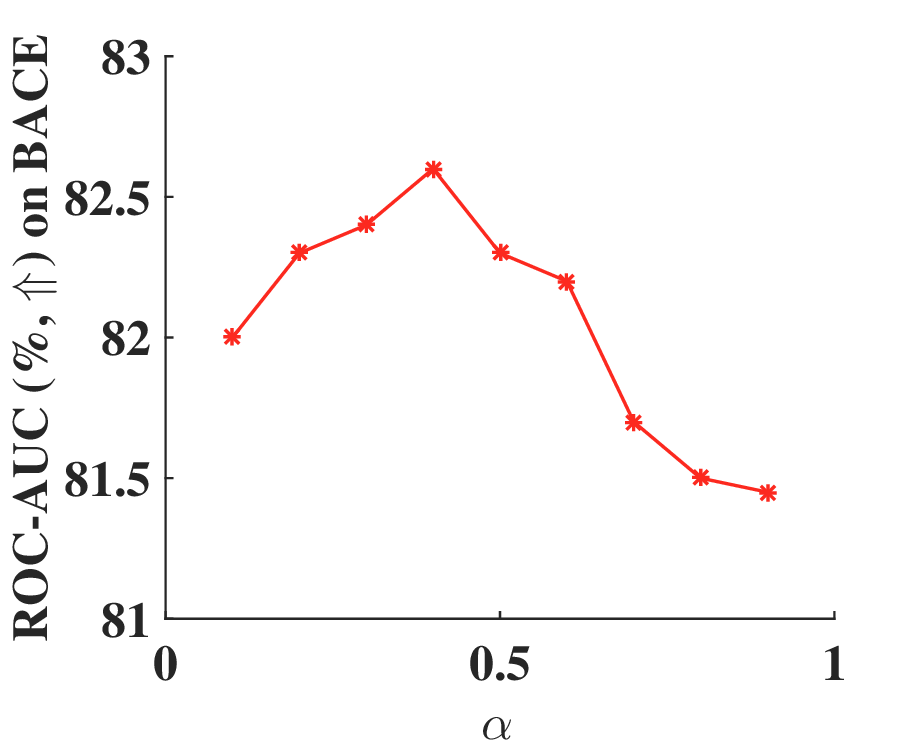}\hfill
    \caption{Performance (ROC-AUC \%, $\Uparrow$) on the BACE task under varying values of $\lambda$ and $\alpha$.}
    \label{fig6}
\end{figure}

\subsection{Verification of Versatility}
To validate the versatility of MMSA, we integrate the proposed multi-modal molecular representation learning and structure-awareness modules into five baseline models. Notably, for the unimodal methods, we remove the cross-modal component from the reconstruction loss \(\mathcal{L}_{rl}\). The main results are shown in Fig. \ref{fig4}. All baselines achieve consistent performance improvements when equipped with the MMSA. This suggests that MMSA has the potential to serve as a universal plugin capable of enhancing any graph-based model (whether 2D or 3D graphs).

\subsection{Influence of GNNs Backbone}
To validate the effectiveness of MMSA across different GNN architectures, we evaluate five distinct GNN models: GCN \cite{kipf2016semi}, GIN \cite{xu2018powerful}, GAT \cite{velickovic2017graph}, GraphSAGE \cite{hamilton2017inductive}, and Graphformer \cite{ying2021transformers}. Table \ref{tab7} presents the performance of these models on four classification and regression datasets, demonstrating that MMSA is architecture-agnostic. Furthermore, we observe that pre-training with GIN and Graphformer yields superior performance. Given that Graphformer has a higher complexity than GIN, we use GIN as the backbone network for downstream tasks.

\subsection{Hyperparametric Analysis}
\begin{table}[t]
\renewcommand{\arraystretch}{1.15}
  \centering
  \caption{Comparison of pre-training performance (ROC-AUC \%, $\Uparrow$ and RMSE, $\Downarrow$) with different $K$ values.}
  \label{tab8}
\begin{adjustbox}{max width=\linewidth}
\begin{tabular}{ccccc}
\toprule
$K$    & BBBP ($\Uparrow$) & Tox21 ($\Uparrow$) & Lipo ($\Downarrow$)  & ESOL ($\Downarrow$) \\ \midrule
2         & 73.8 & 76.0  & 0.710 & 1.064 \\
5         & 74.2 & 77.8  & 0.706 & 1.040 \\
10         & 74.5 & 77.6  & 0.704 & 1.042 \\
15   & 74.2 & 76.7  & 0.707  & 1.045 \\
\bottomrule
\end{tabular}
\end{adjustbox}
\vspace{-1em}
\end{table}

The proposed MMSA framework involves three key hyperparameters: (1) the cross-modal alignment weight $\lambda$, (2) the memory update coefficient $\alpha$, and (3) the neighborhood size $K$ used in hypergraph construction. While our main experiments employ empirically validated settings ($\lambda$ = 0.7, $\alpha$ = 0.2, $K$ = 10), we also conduct a comprehensive sensitivity analysis to evaluate their impact. As illustrated in Fig. \ref{fig6}, the model maintains robust performance within the range of $\lambda \in [0.5, 0.9]$, with the best ROC-AUC score of 82.8 on the BACE task achieved at $\lambda=0.6$, indicating a well-balanced integration of multi-modal information. The behavior of $\alpha$ is more nonlinear: values below 0.2 lead to slow convergence of the memory module, while values above 0.5 induce instability and oscillations during training. As shown in TABLE \ref{tab8}, the influence of $K$ exhibits a quasi-convex trend—small neighborhoods ($K<5$) fail to capture sufficient global structure. In contrast, overly large neighborhoods ($K>10$) introduce noisy connections, which in turn degrade overall performance.

\begin{figure}[t]
    \centering
    \includegraphics[width=0.42\linewidth]{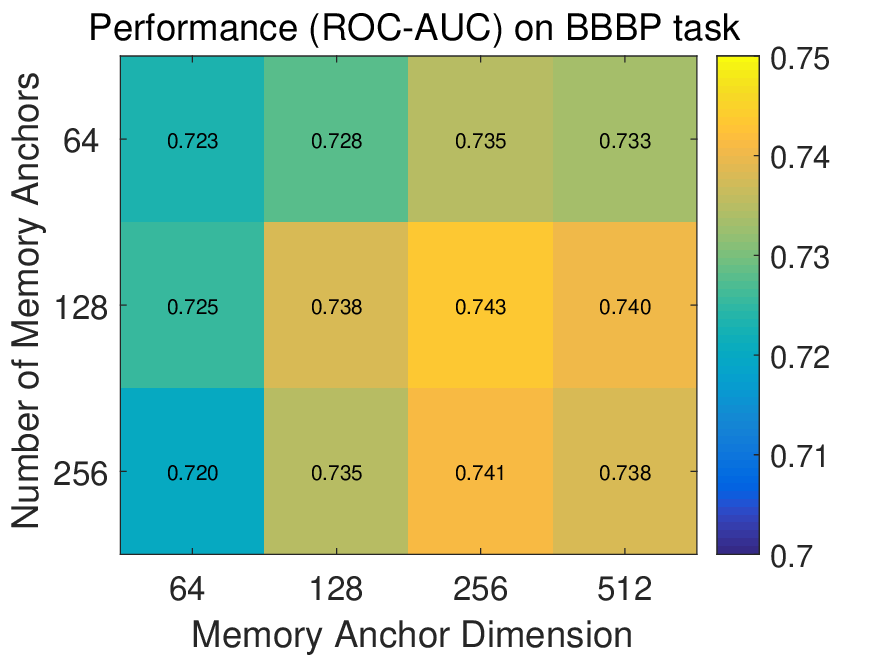}\hfill
    \includegraphics[width=0.42\linewidth]{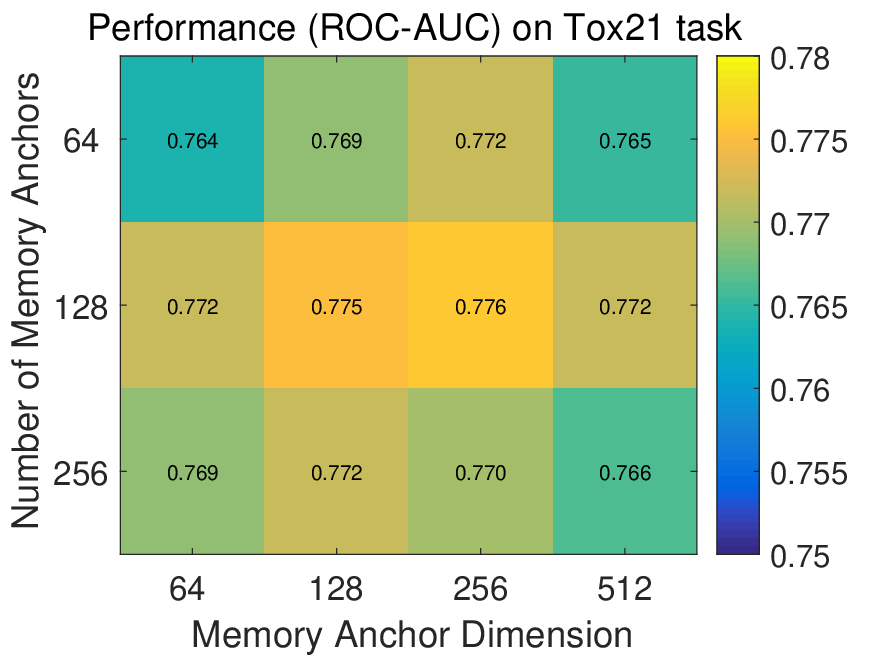}
    \caption{Performance (ROC-AUC) across varying numbers and dimensionalities of memory anchors.}
    \label{fig7}
\vspace{-1em}
\end{figure}

To further investigate the impact of the number and dimensionality of memory anchors on model performance, we conducted a systematic analysis under various configurations (see Fig. \ref{fig7}). The results show that when the number of memory anchors ($L$) is below 64, the model suffers a significant performance drop due to insufficient coverage of the diverse molecular representations. As $L$ increases beyond 128, performance improvements begin to saturate while computational overhead rises sharply. Regarding the dimensionality of memory anchors ($d_c$), values below 128 restrict the model’s ability to capture complex spatial and conformational features of molecules. In contrast, dimensions exceeding 256 tend to introduce overfitting. Balancing efficiency and effectiveness, we select $L=128$ and $d_c=256$ as the final configuration, which achieves optimal performance while maintaining a lightweight model architecture.

\subsection{Visualization}
We visualize the fine-tuned BBBP molecular representations using t-SNE, with colors representing the true labels from the downstream tasks. The clustering results are shown in Fig. \ref{fig5}. Additionally, we quantitatively assess the clustering quality by calculating the Davies-Bouldin index (DBI) \cite{davies1979cluster} and Normalized Mutual Information (NMI). A lower Davies-Bouldin index indicates better clustering performance, while a higher NMI value signifies more precise clustering. For clarity, these values are displayed beneath each corresponding subplot. The experimental results demonstrate that the proposed method generates meaningful clustering outcomes, with samples having the same or similar labels grouped, further validating the effectiveness of MMSA in capturing discriminative molecular features.

\section{Conclusion}
\label{conclusion}
In this paper, we propose a structure-awareness-based multi-modal self-supervised pre-training framework (MMSA) designed to enhance molecular graph representations by leveraging invariant molecular knowledge across different molecules. The framework consists of two key modules: multi-modal molecular representation learning and structure-awareness. The multi-modal molecular representation learning module integrates information from various modalities to generate rich, informative molecular embeddings, while the structure-awareness module captures higher-order interactions and dependencies within molecular structures, enabling a more precise representation of the complex relationships between molecules. MMSA can be seamlessly incorporated into existing 2D/3D graph-based models, resulting in significant performance improvements for molecular representation learning. Our approach achieves state-of-the-art results across multiple classification and regression benchmarks in drug discovery, demonstrating its effectiveness in real-world applications. Furthermore, we validate MMSA's cross-dataset retrieval capabilities, showing its robustness in identifying structurally similar molecules across different datasets. Looking ahead, we aim to extend this framework to handle more complex molecular data and tasks, broadening its applicability to diverse fields in computational biology and drug development.

\ifCLASSOPTIONcompsoc
  \section*{Acknowledgments}
\else
  \section*{Acknowledgment}
\fi
This work is supported in part by the National Natural Science Foundation of China (No.62106259, No.62076234), Beijing Outstanding Young Scientist Program (NO.BJJWZYJH012019100020098), and Beijing Natural Science Foundation (No. 4222029).

\section*{Reproducibility Statement}
We are committed to ensuring the reproducibility of the study. All code, datasets, and model configurations used in this study are publicly available. The code is accessible on \href{https://1drv.ms/f/c/c7329276c9d62249/En4sf_eXLN9Gil1qxprw4JkBSS14XY-jvQAvyhH9J9P_Ew?e=dKM2Lm}{here}, where we provide detailed instructions for pre-processing data and running experiments.

\ifCLASSOPTIONcaptionsoff
  \newpage
\fi

\bibliographystyle{IEEEtran}
\bibliography{bare_jrnl_new_sample4}
\newpage

\begin{IEEEbiography}[{\includegraphics[width=1in,height=1.25in,clip,keepaspectratio]{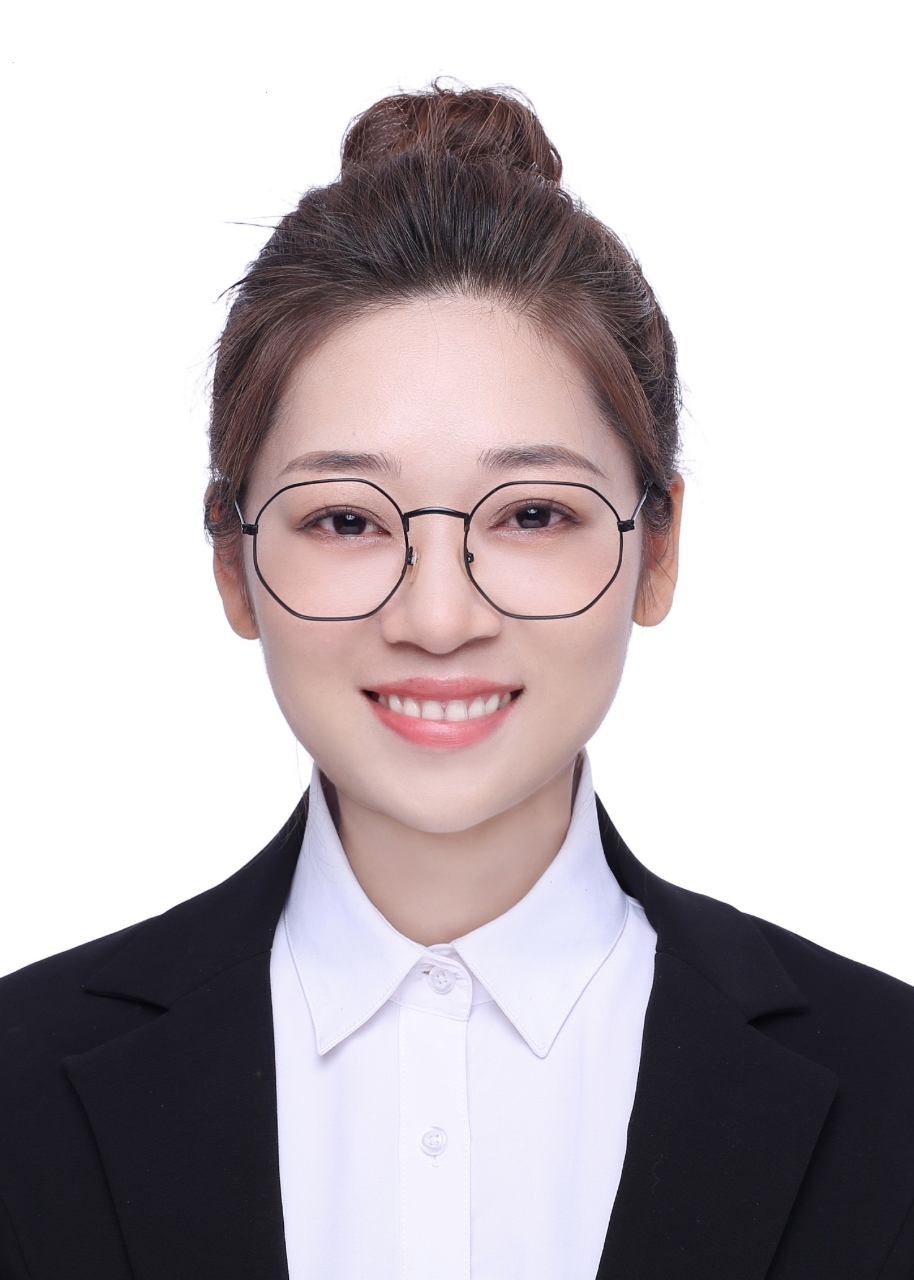}}]{Rong Yin} received the Ph.D. degree from the School of Cyber Security, University of Chinese Academy of Sciences, Beijing, China, in 2020. She is currently an Associate Professor with the Institute of Information Engineering, Chinese Academy of Sciences, Beijing, China. Her current research interests include large-scale machine learning, graph representation learning, statistical machine learning theory, etc. She has published more than 20 papers on top-tier conferences and journals in artificial intelligence, e.g., NeurIPS, ICML, IJCAI, AAAI, IEEE Transactions on Image Processing, IEEE Transactions on Knowledge and Data Engineering, IEEE Transactions on Neural Networks and Learning Systems, IEEE Transaction on Multimedia, Information Fusion, etc. She received the “Excellent Talent Introduction” of Institute of Information Engineering, CAS. She served as the program committee of several conferences, e.g., NeurIPS, ICML, ICLR, AAAI etc.
\end{IEEEbiography}
\vspace{-1em}

\begin{IEEEbiography}[{\includegraphics[width=1in,height=1.25in,clip,keepaspectratio]{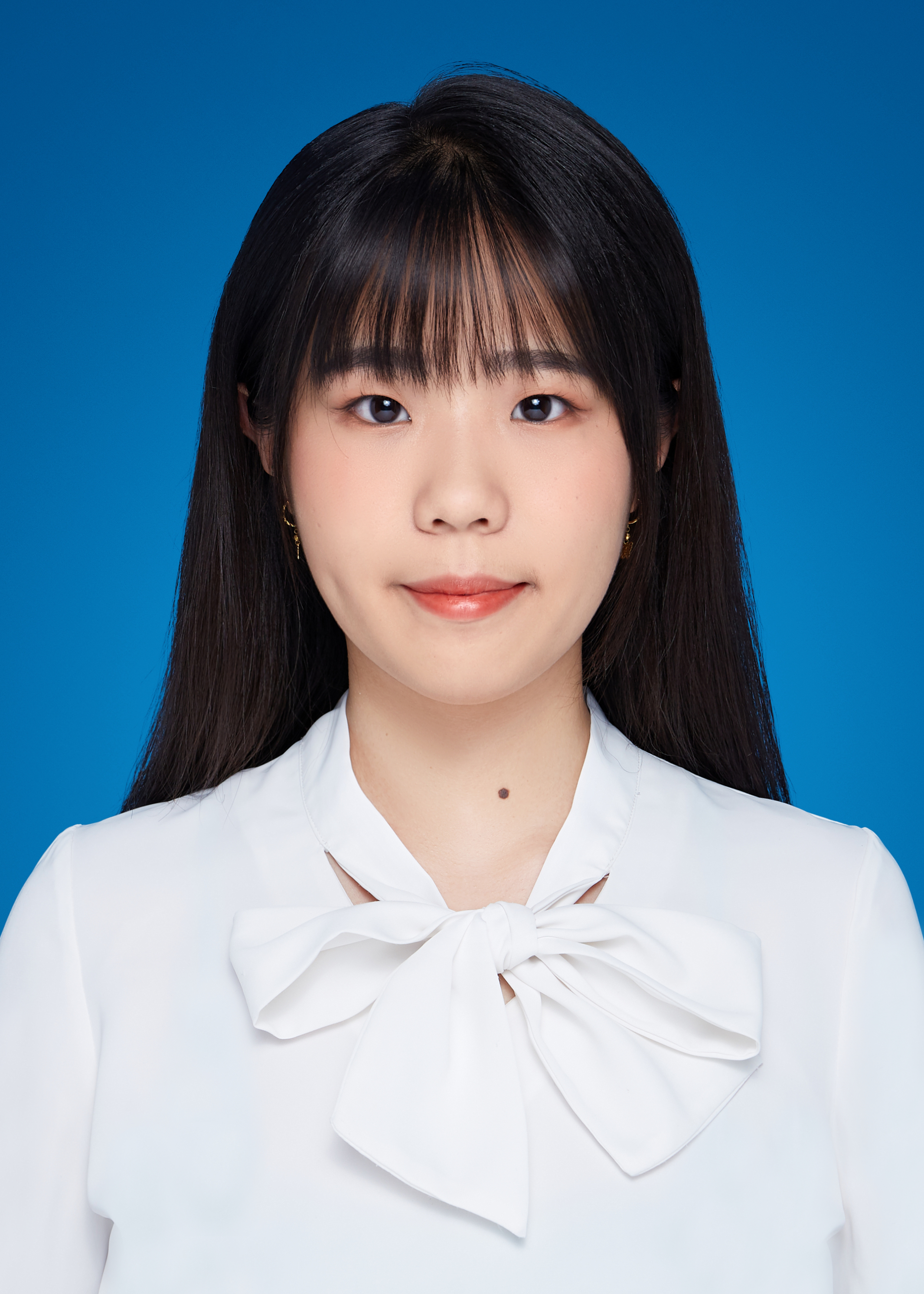}}]{Ruyue Liu}
is a Ph.D. student at the Institute of Information Engineering, Chinese Academy of Sciences, Beijing, China, and the School of Cyber Security, University of Chinese Academy of Sciences, Beijing, China. Her current research interests include machine learning, data mining, self-supervised learning, federated learning, and graph representation learning.
\end{IEEEbiography}
\vspace{-1em}

\begin{IEEEbiography}[{\includegraphics[width=1in,height=1.25in,clip,keepaspectratio]{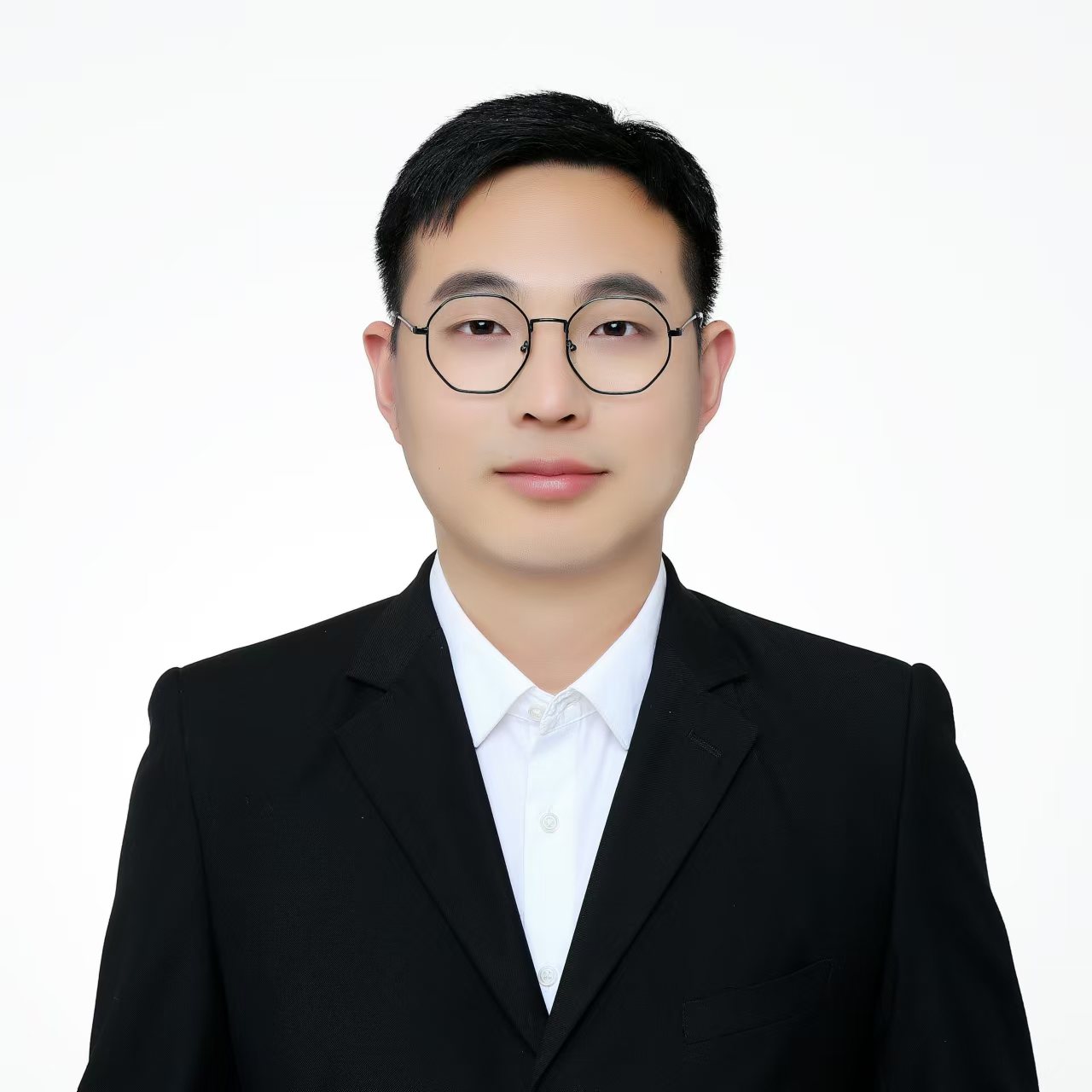}}]{Xiaoshuai Hao} obtained his Ph.D. from the Institute of Information Engineering at the Chinese Academy of Sciences in 2023. He is currently a researcher specializing in embodied multimodal large models at the Beijing Academy of Artificial Intelligence. His research interests encompass multimedia retrieval, multimodal learning, and embodied intelligence.
\end{IEEEbiography}
\vspace{-1em}

\begin{IEEEbiography}[{\includegraphics[width=1in,height=1.25in,clip,keepaspectratio]{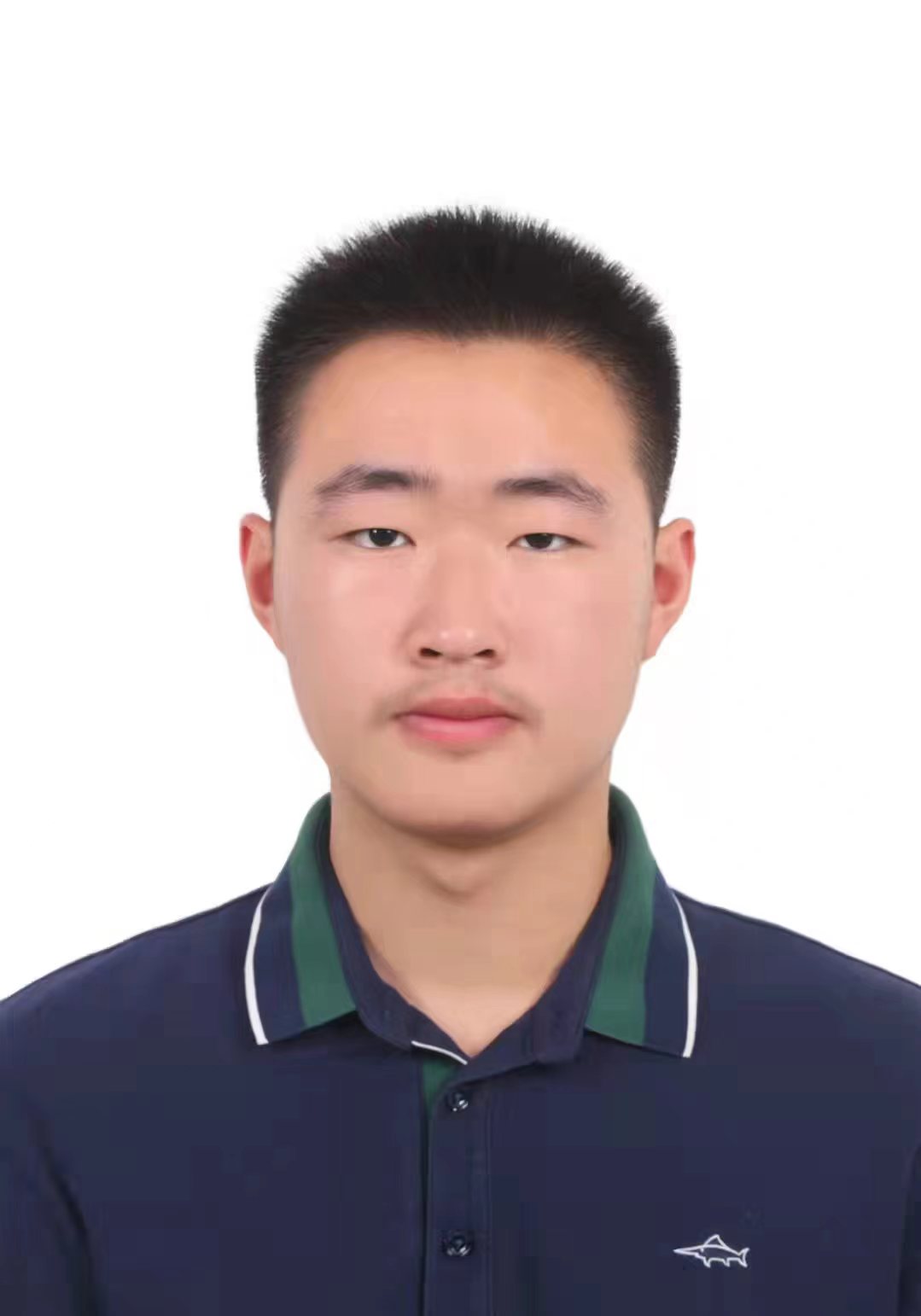}}]{Xingrui Zhou} is an undergraduate at Xidian University, Xi'an, China. His current research interests include machine learning, numerical analysis, optimization algorithms, and computational electromagnetics.
\end{IEEEbiography}
\vspace{-1em}

\begin{IEEEbiography}[{\includegraphics[width=1in,height=1.25in,clip,keepaspectratio]{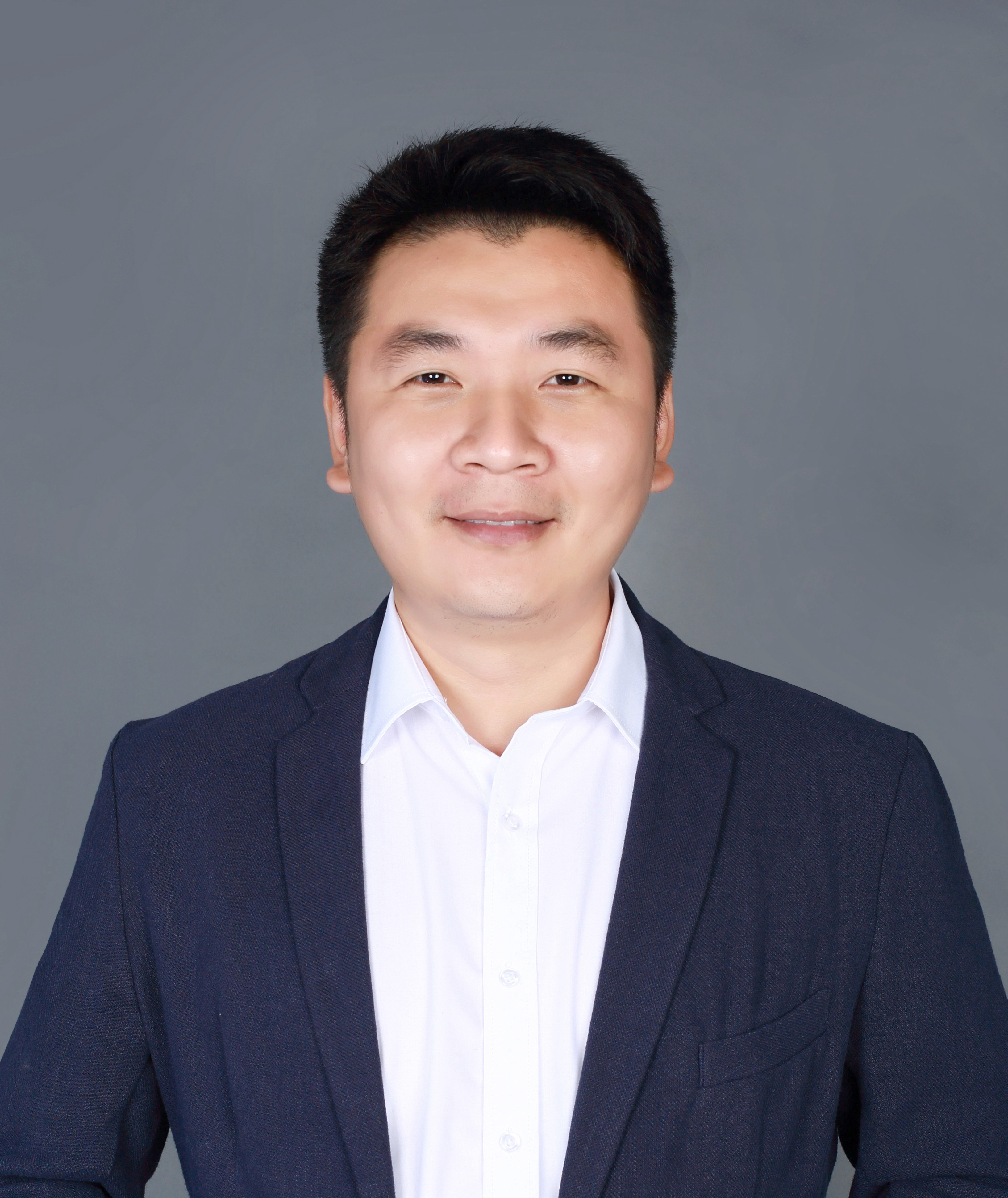}}]{Yong Liu}
received the PhD degree in computer science from Tianjin University, Tianjin, China, in 2016. He is tenure-track faculty with the Gaoling School of Artificial Intelligence, Renmin University of China. His research interests are mainly about machine learning, with special attention to large-scale machine learning, autoML, statistical machine learning theory, etc. He has published more than 30 papers on top-tier conferences and journals in artificial intelligence, e.g., IEEE Transactions on Pattern Analysis and Machine Intelligence, NeurIPS, ICML, IJCAI, AAAI, IEEE Transactions on Image Processing, IEEE Transactions on Neural Networks and Learning Systems, etc. He received the ”Youth Innovation Promotion Association” of CAS and the “Excellent Talent Introduction” of Institute of Information Engineering, CAS. He served as the program committee of several
conferences, e.g., NeurIPS, AAAI, IJCAI, ECAI etc.
\end{IEEEbiography}
\vspace{-1em}

\begin{IEEEbiography}[{\includegraphics[width=1in,height=1.25in,clip,keepaspectratio]{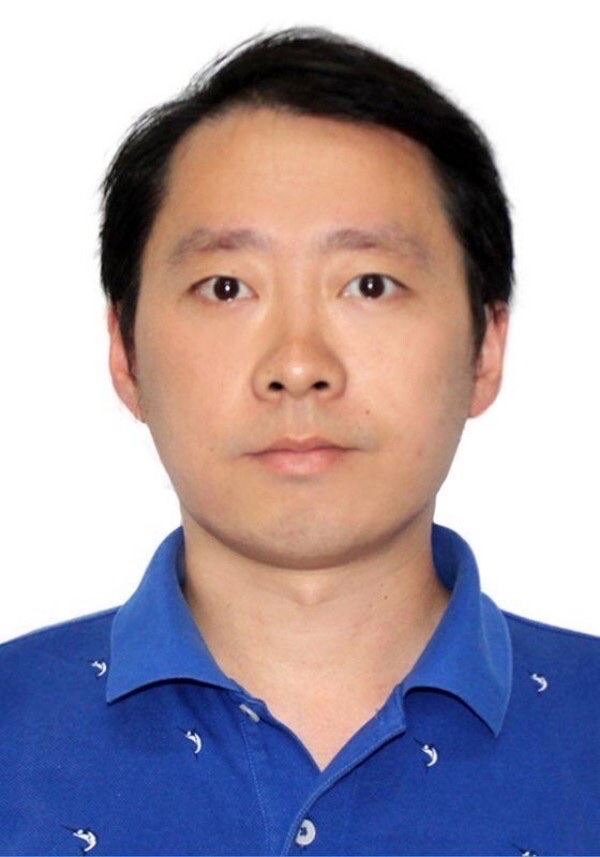}}]{Can Ma}
received his Ph.D. degree from the Institute of Computing Technology, Chinese Academy of Sciences in 2012. He is currently a full senior engineer and PhD supervisor at the Institute of Information Engineering, Chinese Academy of Sciences. His research interests include big data management and analysis in cyberspace and social network analysis.
\end{IEEEbiography} 
\vspace{-45em}

\begin{IEEEbiography}[{\includegraphics[width=1in,height=1.25in,clip,keepaspectratio]{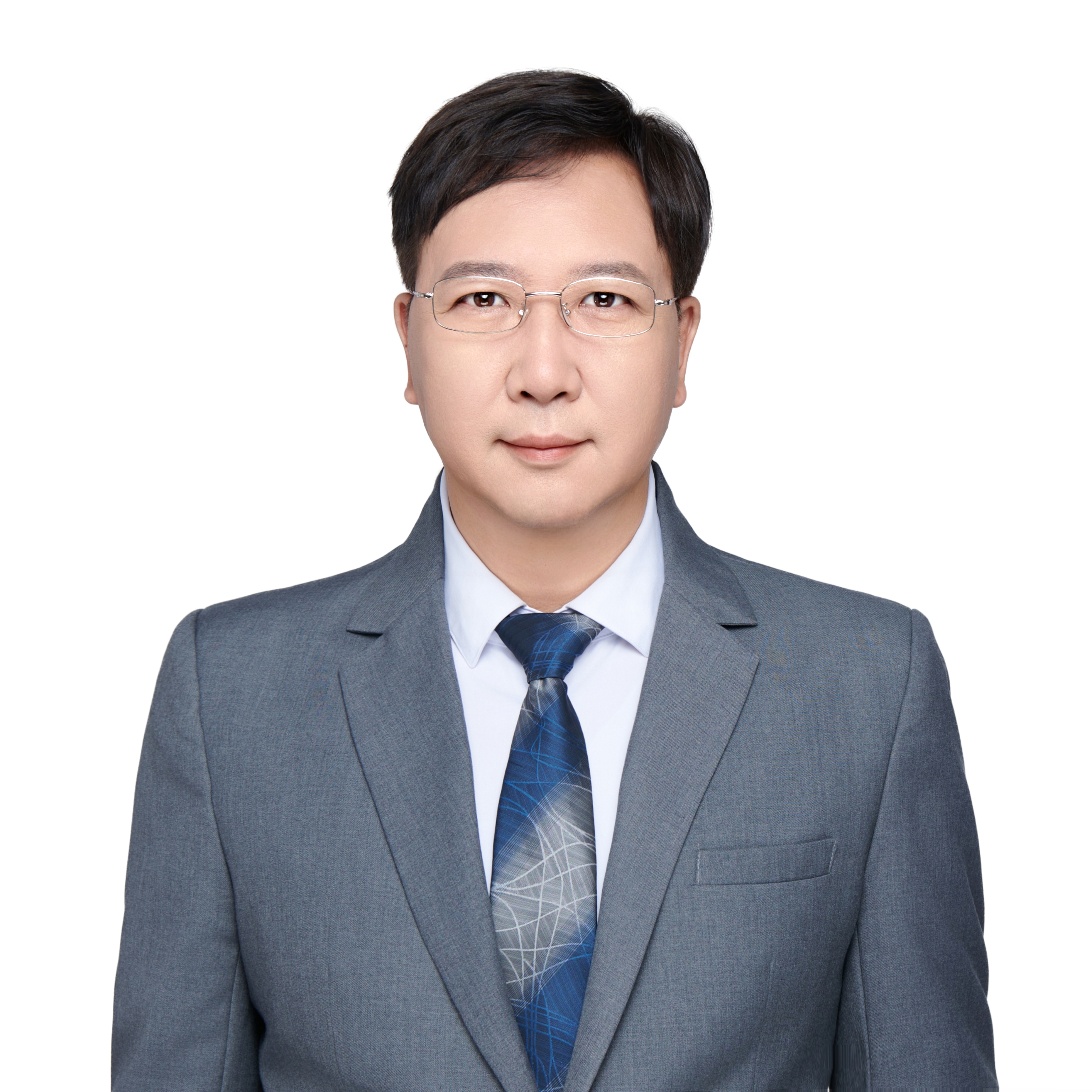}}]{Weiping Wang}
 received a Ph.D. degree in computer science from the Harbin Institute of Technology, Harbin, China, in 2006. He is currently a Professor at the Institute of Information Engineering, Chinese Academy of Sciences, Beijing, China. His research interests include big data, data security, databases, and storage systems. He has more than 100 publications in major journals and international conferences.
\end{IEEEbiography}

\end{document}